\documentclass[11pt,a4paper]{article}
\usepackage[hyperref]{acl2021}
\usepackage{times}
\usepackage{latexsym}

\usepackage{amsmath,amsfonts,bm}

\def\eqref#1{equation~\ref{#1}}

\def\1{\bm{1}}

\def\vtheta{{\bm{\theta}}}
\def\va{{\bm{a}}}
\def\vb{{\bm{b}}}

\def\vr{{\bm{r}}}

\def\vv{{\bm{v}}}

\def\vx{{\bm{x}}}
\def\vy{{\bm{y}}}

\def\eva{{a}}

\def\evr{{r}}

\def\mV{{\bm{V}}}
\def\mW{{\bm{W}}}

\DeclareMathAlphabet{\mathsfit}{\encodingdefault}{\sfdefault}{m}{sl}
\SetMathAlphabet{\mathsfit}{bold}{\encodingdefault}{\sfdefault}{bx}{n}

\def\sD{{\mathbb{D}}}

\def\sY{{\mathbb{Y}}}

\newcommand{\E}{\mathbb{E}}

\newcommand{\R}{\mathbb{R}}

\usepackage{url}
\usepackage{hyperref}
\usepackage{url}
\usepackage{bbm}
\usepackage{graphicx}
\usepackage{amsmath}
\usepackage{amsfonts}
\usepackage{boldline}
\usepackage{caption}
\usepackage{tikz-dependency}
\usepackage{wrapfig}
\usepackage{multirow}
\usepackage{pgfplots}
\usepackage{filecontents}
\usepackage[T1]{fontenc}
\usepackage{subfiles}
\usepackage{readarray}
\usepackage{xcolor}
\usepackage{import}
\usepackage{hhline}
\usepackage{mathtools}
\usepackage{multicol}
\usepackage{enumitem}

\newcommand{\Score}{\text{Score}}

\usepackage{microtype}

\aclfinalcopy 
\def\aclpaperid{444} 

\title{Automated Concatenation of Embeddings for Structured Prediction}

\author{Xinyu Wang$^{\diamond\ddagger}$, Yong Jiang$^{\dagger}$\textsuperscript{$\ast$}, Nguyen Bach$^{\dagger}$, Tao Wang$^{\dagger}$,\\
\textbf{Zhongqiang Huang$^{\dagger}$, Fei Huang$^{\dagger}$,  Kewei Tu$^{\diamond}$}\thanks{\hspace{1mm} Yong Jiang and Kewei Tu are the corresponding authors. $^{\ddagger}$: This work was conducted when Xinyu Wang was interning at Alibaba DAMO Academy. } \\
 $^\diamond$School of Information Science and Technology, ShanghaiTech University \\
 Shanghai Engineering Research Center of Intelligent Vision and Imaging \\
  Shanghai Institute of Microsystem and Information Technology, Chinese Academy of Sciences \\
 University of Chinese Academy of Sciences \\
 $^\dagger$DAMO Academy, Alibaba Group \\
  {\tt \{wangxy1,tukw\}@shanghaitech.edu.cn, yongjiang.jy@alibaba-inc.com} \\
  {\tt \{nguyen.bach,leeo.wangt,z.huang,f.huang\}@alibaba-inc.com} \\
}

\date{}

\begin{document}
\maketitle

\begin{abstract}
Pretrained contextualized embeddings are powerful word representations for structured prediction tasks. Recent work found that better word representations can be obtained by concatenating different types of embeddings. However, the selection of embeddings to form the best concatenated representation usually varies depending on the task and the collection of candidate embeddings, and the ever-increasing number of embedding types makes it a more difficult problem. In this paper, we propose Automated Concatenation of Embeddings (ACE) to automate the process of finding better concatenations of embeddings for structured prediction tasks, based on a formulation inspired by recent progress on neural architecture search. Specifically, a controller alternately samples a concatenation of embeddings, according to its current belief of the effectiveness of individual embedding types in consideration for a task, and updates the belief based on a reward. We follow strategies in reinforcement learning to optimize the parameters of the controller and compute the reward based on the accuracy of a task model, which is fed with the sampled concatenation as input and trained on a task dataset. Empirical results on 6 tasks and 21 datasets show that our approach outperforms strong baselines and achieves state-of-the-art performance with fine-tuned embeddings in all the evaluations.\footnote{Our code is publicly available at \url{https://github.com/Alibaba-NLP/ACE}.}

\end{abstract}

\section{Introduction}

Recent developments on pretrained contextualized embeddings have significantly improved the performance of structured prediction tasks in natural language processing. Approaches based on contextualized embeddings, such as ELMo \citep{peters-etal-2018-deep}, Flair \citep{akbik-etal-2018-contextual}, BERT \citep{devlin-etal-2019-bert}, and XLM-R \citep{conneau-etal-2020-unsupervised}, have been consistently raising the state-of-the-art for various structured prediction tasks. Concurrently, research has also showed that word representations based on the concatenation of multiple pretrained contextualized embeddings and traditional non-contextualized embeddings (such as word2vec \citep{mikolov2013distributed} and character embeddings \citep{santos2014learning}) can further improve performance \citep{peters-etal-2018-deep,akbik-etal-2018-contextual,strakova-etal-2019-neural,wang-etal-2020-more}. Given the ever-increasing number of embedding learning methods that operate on different granularities (e.g., word, subword, or character level) and with different model architectures, choosing the best embeddings to concatenate for a specific task becomes non-trivial, and exploring all possible concatenations can be prohibitively demanding in computing resources.


Neural architecture search (NAS) is an active area of research in deep learning to automatically search for better model architectures, and has achieved state-of-the-art performance on various tasks in computer vision, such as image classification \citep{real2019regularized}, semantic segmentation \citep{liu2019auto}, and object detection \citep{ghiasi2019fpn}. In natural language processing, NAS has been successfully applied to find better RNN structures  \citep{zoph2016neural,pham2018efficient} and recently better transformer structures \citep{so2019evolved,zhu2020autotrans}. In this paper, we propose Automated Concatenation of Embeddings (ACE) to automate the process of finding better concatenations of embeddings for structured prediction tasks. ACE is formulated as an NAS problem. In this approach, an iterative search process is guided by a controller based on its belief that models the effectiveness of individual embedding candidates in consideration for a specific task. At each step, the controller samples a concatenation of embeddings according to the belief model and then feeds the concatenated word representations as inputs to a task model, which in turn is trained on the task dataset and returns the model accuracy as a reward signal to update the belief model. We use the policy gradient algorithm \citep{williams1992simple} in reinforcement learning \citep{sutton2018reinforcement} to solve the optimization problem. In order to improve the efficiency of the search process, we also design a special reward function by accumulating all the rewards based on the transformation between the current concatenation and all previously sampled concatenations.

Our approach is different from previous work on NAS in the following aspects:
\begin{enumerate}[leftmargin=*]
	\item Unlike most previous work, we focus on searching for better word representations rather than better model architectures.
    \item We design a novel search space for the embedding concatenation search. Instead of using RNN as in previous work of \citet{zoph2016neural}, we design a more straightforward controller to generate the embedding concatenation. We design a novel reward function in the objective of optimization to better evaluate the effectiveness of each concatenated embeddings.
    \item ACE achieves high accuracy without the need for retraining the task model, which is typically required in other NAS approaches.
    \item Our approach is efficient and practical. Although ACE is formulated in a NAS framework, ACE can find a strong word representation on a single GPU with only a few GPU-hours for structured prediction tasks. In comparison, a lot of NAS approaches require dozens or even thousands of GPU-hours to search for good neural architectures for their corresponding tasks.
\end{enumerate}

Empirical results show that ACE outperforms strong baselines. 
Furthermore, when ACE is applied to concatenate pretrained contextualized embeddings fine-tuned on specific tasks, we can achieve
state-of-the-art accuracy on 6 structured prediction tasks including Named Entity Recognition \citep{Sundheim1995NamedET}, Part-Of-Speech tagging \citep{derose-1988-grammatical}, chunking \citep{tjong-kim-sang-buchholz-2000-introduction}, aspect extraction \citep{10.1145/1014052.1014073}, syntactic dependency parsing \citep{arrive1969elements} and semantic dependency parsing \citep{oepensemeval} over 21 datasets.
Besides, we also analyze the advantage of ACE and reward function design over the baselines and show the advantage of ACE over ensemble models. 

\section{Related Work}
\subsection{Embeddings}


Non-contextualized embeddings, such as word2vec \citep{mikolov2013distributed}, GloVe \citep{pennington2014glove}, and fastText \citep{bojanowski2017enriching}, help lots of NLP tasks. 
Character embeddings \citep{santos2014learning} are trained together with the task and applied in many structured prediction tasks \citep{ma-hovy-2016-end,lample-etal-2016-neural,dozat-manning-2018-simpler}. 
For pretrained contextualized embeddings, ELMo \citep{peters-etal-2018-deep}, a pretrained contextualized word embedding generated with multiple Bidirectional LSTM layers, significantly outperforms previous state-of-the-art approaches on several NLP tasks. 
Following this idea, \citet{akbik-etal-2018-contextual} proposed Flair embeddings, which is a kind of contextualized character embeddings and achieved strong performance in sequence labeling tasks. 
Recently, \citet{devlin-etal-2019-bert} proposed BERT, which encodes contextualized sub-word information by Transformers \citep{vaswani2017attention} and significantly improves the performance on a lot of NLP tasks. 
Much research such as RoBERTa \citep{liu2019roberta} has focused on improving BERT model's performance through stronger masking strategies. Moreover, multilingual contextualized embeddings become popular. \citet{pires-etal-2019-multilingual} and \citet{wu-dredze-2019-beto} showed that Multilingual BERT (M-BERT) could learn a good multilingual representation effectively with strong cross-lingual zero-shot transfer performance in various tasks. \citet{conneau-etal-2020-unsupervised} proposed XLM-R, which is trained on a larger multilingual corpus and significantly outperforms M-BERT on various multilingual tasks.

\subsection{Neural Architecture Search}
Recent progress on deep learning has shown that network architecture design is crucial to the model performance. 
However, designing a strong neural architecture for each task requires enormous efforts, high level of knowledge, and experiences over the task domain. 
Therefore, automatic design of neural architecture is desired. 
A crucial part of NAS is search space design, which defines the discoverable NAS space. 
Previous work \citep{baker2016designing,zoph2016neural,xie2017genetic} designs a global search space \citep{elsken2019neural} which incorporates structures from hand-crafted architectures. For example, \citet{zoph2016neural} designed a chained-structured search space with skip connections. 
The global search space usually has a considerable degree of freedom. 
For example, the approach of \citet{zoph2016neural} takes 22,400 GPU-hours to search on CIFAR-10 dataset. Based on the observation that existing hand-crafted architectures contain repeated structures \citep{szegedy2016rethinking,he2016deep,huang2017densely}, \citet{zoph2018learning} explored cell-based search space which can reduce the search time to 2,000 GPU-hours. 

In recent NAS research, reinforcement learning and evolutionary algorithms are the most usual approaches. 
In reinforcement learning, the agent's actions are the generation of neural architectures and the action space is identical to the search space. Previous work usually applies an RNN layer \citep{zoph2016neural,zhong2018practical,zoph2018learning} or use Markov Decision Process \citep{baker2016designing} to decide the hyper-parameter of each structure and decide the input order of each structure.
Evolutionary algorithms have been applied to architecture search for many decades \citep{geoffrey1989designing,angeline1994evolutionary,stanley2002evolving,floreano2008neuroevolution,jozefowicz2015empirical}. The algorithm repeatedly generates new populations through recombination and mutation operations and selects survivors through competing among the population. Recent work with evolutionary algorithms differ in the method on parent/survivor selection and population generation. For example, \citet{real2017large}, \citet{liu2018hierarchical}, \citet{wistuba2018deep} and \citet{real2019regularized} applied tournament selection \citep{goldberg1991comparative} for the parent selection while \citet{xie2017genetic} keeps all parents. \citet{suganuma2017genetic} and \citet{elsken2017simple} chose the best model while \citet{real2019regularized} chose several latest models as survivors.



\section{Automated Concatenation of Embeddings}
\begin{figure*}
	\centering
	\includegraphics[scale=0.8]{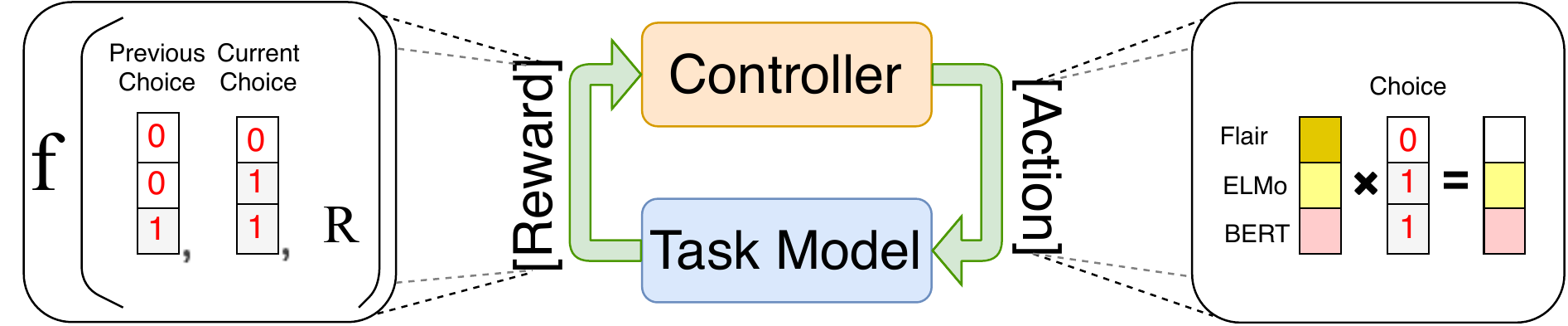}
	\caption{The main paradigm of our approach is shown in the middle, where an example of reward function is represented in the left and an example of a concatenation action is shown in the right.}
	\label{fig:architecture}
\end{figure*}

In ACE, a task model and a controller interact with each other repeatedly. The task model predicts the task output, while the controller searches for better embedding concatenation as the word representation for the task model to achieve higher accuracy. Given an embedding concatenation generated from the controller, the task model is trained over the task data and returns a reward to the controller. The controller receives the reward to update its parameter and samples a new embedding concatenation for the task model. 
Figure \ref{fig:architecture} shows the general architecture of our approach.
\subsection{Task Model}
For the task model, we emphasis on sequence-structured and graph-structured outputs. Given a structured prediction task with input sentence $\vx$ and structured output $\vy$, we can calculate the probability distribution $P(\vy|\vx)$ by:
\begin{align}
P(\vy|\vx) = \frac{\exp{(\Score(\vx,\vy)})}{\sum_{\vy^{\prime} \in \sY(\vx)} \exp{(\Score(\vx,\vy^{\prime})})} \nonumber
\end{align}
where $\sY(\vx)$ represents all possible output structures given the input sentence $\vx$. Depending on different structured prediction tasks, the output structure $\vy$ can be label sequences, trees, graphs or other structures. In this paper, we use sequence-structured and graph-structured outputs as two exemplar structured prediction tasks. We use BiLSTM-CRF model \citep{ma-hovy-2016-end,lample-etal-2016-neural} for sequence-structured outputs and use BiLSTM-Biaffine model \citep{dozat2016deep} for graph-structured outputs:
\begin{align}
P^{\text{seq}}(\vy|\vx) &= \text{BiLSTM-CRF}(\mV,\vy)\nonumber\\
P^{\text{graph}}(\vy|\vx) &= \text{BiLSTM-Biaffine}(\mV,\vy)\nonumber
\end{align}
where $\mV = [\vv_1; \cdots; \vv_n]$, $\mV \in \R^{d\times n}$ is a matrix of the word representations for the input sentence $\vx$ with $n$ words, $d$ is the hidden size of the concatenation of all embeddings. The word representation $\vv_i$ of $i$-th word is a concatenation of $L$ types of word embeddings:
\begin{align}
\vv_i^l &= \text{embed}_i^l (\vx); \;\; \vv_i = [\vv_i^1;\vv_i^2; \dots; \vv_i^L] \nonumber
\end{align}
where $\text{embed}^l$ is the model of $l$-th embeddings, $\vv_i\in \R^d$, $\vv_i^l\in \R^{d^l}$. $d^l$ is the hidden size of $\text{embed}^l$. 

\subsection{Search Space Design}
\label{sec:search_space}
The neural architecture search space can be represented as a set of neural networks \citep{elsken2019neural}. 
A neural network can be represented as a directed acyclic graph with a set of nodes and directed edges.
Each node represents an operation, while each edge represents the inputs and outputs between these nodes. 
In ACE, we represent each embedding candidate as a node. 
The input to the nodes is the input sentence $\vx$, and the outputs are the embeddings $\vv^l$. Since we concatenate the embeddings as the word representation of the task model, there is no connection between nodes in our search space. Therefore, the search space can be significantly reduced. 
For each node, there are a lot of options to extract word features. 
Taking BERT embeddings as an example, \citet{devlin-etal-2019-bert} concatenated the last four layers as word features while \citet{kondratyuk-straka-2019-75} applied a weighted sum of all twelve layers. However, the empirical results \citep{devlin-etal-2019-bert} do not show a significant difference in accuracy. We follow the typical usage for each embedding to further reduce the search space. As a result, each embedding only has a fixed operation and the resulting search space contains $2^L{-}1$ possible combinations of nodes.


In NAS, weight sharing \citep{pmlr-v80-pham18a} shares the weight of structures in training different neural architectures to reduce the training cost. In comparison, we fixed the weight of pretrained embedding candidates in ACE except for the character embeddings. 
Instead of sharing the parameters of the embeddings, we share the parameters of the task models at each step of search.
However, the hidden size of word representation varies over the concatenations, making the weight sharing of structured prediction models difficult. Instead of deciding whether each node exists in the graph, we keep all nodes in the search space and add an additional operation for each node to indicate whether the embedding is masked out. To represent the selected concatenation, we use a binary vector $\va=[a_1, \cdots, a_l, \cdots, a_L]$ as an mask to mask out the embeddings which are not selected:
\begin{align}
\vv_i = [\vv_i^1a_1;\dots;\vv_i^la_l; \dots; \vv_i^La_L] \label{eq:vector}
\end{align}
where $a_l$ is a binary variable. Since the input $\mV$ is applied to a linear layer in the BiLSTM layer, multiplying the mask with the embeddings is equivalent to directly concatenating the selected embeddings:
\begin{align}
\mW^{\top}\vv_i = \sum_{l=1}^L \mW_l^{\top}\vv_i^l a_l \label{eq:linear}
\end{align}
where $\mW {=} [\mW_1;\mW_2;\dots ;\mW_L]$ and $\mW{\in} \R ^ {d \times h}$ and $\mW_l{\in} \R ^ {d^l \times h}$. Therefore, the model weights can be shared after applying the embedding mask to all embedding candidates' concatenation. Another benefit of our search space design is that we can remove the unused embedding candidates and the corresponding weights in $\mW$ for a lighter task model after the best concatenation is found by ACE.

\subsection{Searching in the Space}
During search, the controller generates the embedding mask for the task model iteratively. 
We use parameters $\vtheta=[\theta_1;\theta_2;\dots;\theta_L]$ for the controller instead of using the RNN structure applied in previous approaches \citep{zoph2016neural,zoph2018learning}. The probability distribution of selecting an concatenation $\va$ is $P^{\text{ctrl}}(\va;\vtheta)=\prod_{l=1}^LP_l^{\text{ctrl}}(a_l;\theta_l)$. Each element $a_l$ of $\va$ is
sampled independently from a Bernoulli distribution,
which is defined as:
\begin{align}
P_l^{\text{ctrl}}(a_l;\theta_l) {=}
    \begin{cases}
    \sigma(\theta_l) &a_l{=}1\\
    1{-} P_l^{\text{ctrl}}(a_l{=}1;\theta_l) &a_l{=}0
    \end{cases}
    \label{eq:prob}
\end{align}
where $\sigma$ is the sigmoid function.
Given the mask, the task model is trained until convergence and returns an accuracy $R$ on the development set. 
As the accuracy cannot be back-propagated to the controller, we use the reinforcement algorithm for optimization. 
The accuracy $R$ is used as the reward signal to train the controller. 
The controller's target is to maximize the expected reward $J(\vtheta)=\E_{P^{\text{ctrl}}(\va;\vtheta)}[R]$ through the policy gradient method \citep{williams1992simple}.
In our approach, since calculating the exact expectation is intractable, the gradient of $J(\vtheta)$ is approximated by sampling only one selection following the distribution $P^{\text{ctrl}}(\va;\vtheta)$ at each step for training efficiency:
\begin{align}
\nabla_\vtheta J(\vtheta) \approx  \sum_{l=1}^L  \nabla_\vtheta \log P_l^{\text{ctrl}}(\eva_l;\theta_l) (R-b) \label{eq:rl}
\end{align} 
where $b$ is the baseline function to reduce the high variance of the update function. The baseline usually can be the highest accuracy during the search process. 
Instead of merely using the highest accuracy of development set over the search process as the baseline, we design a reward function on how each embedding candidate contributes to accuracy change by utilizing all searched concatenations' development scores.
We use a binary vector $|\va^t-\va^{i}|$ to represent the change between current embedding concatenation $\va^t$ at current time step $t$ and $\va^{i}$ at previous time step $i$.
We then define the reward function as:
\begin{align}
\vr^t = \sum_{i=1}^{t-1} (R_t-R_i) |\va^t-\va^{i}| \label{eq:reward_a}
\end{align}
where $\vr^t$ is a vector with length $L$ representing the reward of each embedding candidate. $R_t$ and $R_i$ are the reward at time step $t$ and $i$. 
When the Hamming distance of two concatenations $Hamm(\va^t,\va^{i})$ gets larger, the changed candidates' contribution to the accuracy becomes less noticeable. The controller may be misled to reward a candidate that is not actually helpful.
We apply a discount factor to reduce the reward for two concatenations with a large Hamming distance to alleviate this issue.
Our final reward function is:
\begin{align}
\vr^t {=} \sum_{i=1}^{t-1} (R_t{-}R_i) \gamma ^{Hamm(\va^t,\va^{i})-1} |\va^t{-}\va^{i}| \label{eq:reward}
\end{align}
where $\gamma \in (0,1)$. Eq. \ref{eq:rl} is then reformulated as:
\begin{align}
\nabla_\vtheta J_t(\vtheta) \approx \sum_{l=1}^L \nabla_\vtheta \log P_l^{\text{ctrl}}(\eva_l^t;\theta_l) \evr^t_{l} \label{eq:gradient}
\end{align} 

\subsection{Training}
To train the controller, we use a dictionary $\sD$ to store the concatenations and the corresponding validation scores. At $t=1$, we train the task model with all embedding candidates concatenated. From $t=2$, we repeat the following steps until a maximum iteration $T$:
\begin{enumerate}[leftmargin=*]
    \item Sample a concatenation $\va^t$ based on the probability distribution in Eq. \ref{eq:prob}.
    \item Train the task model with $\va^t$ following Eq. \ref{eq:vector} and evaluate the model on the development set to get the accuracy $R_t$.
    \item Given the concatenation $\va^t$, accuracy $R_t$ and $\sD$, compute the gradient of the controller following Eq. \ref{eq:gradient} and update the parameters of controller.
    \item Add $\va^t$ and $R_t$ into $\sD$, set $t=t+1$.
\end{enumerate}
When sampling $\va^t$, we avoid selecting the previous concatenation $\va^{t-1}$ and the all-zero vector (i.e., selecting no embedding). If $\va^t$ is in the dictionary $\sD$, we compare the $R_t$ with the value in the dictionary and keep the higher one.

\begin{table*}[!ht]
\small
\centering
\setlength\tabcolsep{2.25pt}
\begin{tabular}{l||cccc|ccc|cccccccc}
\hlineB{4}
      & \multicolumn{4}{c|}{\bf \textsc{NER}}     & \multicolumn{3}{c|}{\bf \textsc{POS}}   & \multicolumn{8}{c}{\bf \textsc{AE}}          \\
\hhline{~||----|---|--------}
      & de       & en         & es   & nl  & Ritter & ARK    & TB-v2   & 14Lap  & 14Res   & 15Res   & 16Res & es   & nl   & ru   & tr   \\
\hline\hline
\bf \textsc{All}    & 83.1        & 92.4        & \textbf{88.9} & 89.8  & 90.6   & 92.1   & 94.6    & 82.7   & 88.5    & 74.2    & 73.2  & 74.6 & 75.0 & 67.1 & 67.5 \\
\bf \textsc{Random} & 84.0        & 92.6        & 88.8 & 91.9  & 91.3   & 92.6   & 94.6    & 83.6   & 88.1    & 73.5    & 74.7  & 75.0 & 73.6 & 68.0 & 70.0 \\
\bf \textsc{ACE}    & \textbf{84.2}        & \textbf{93.0}        & \textbf{88.9} & \textbf{92.1}  & \textbf{91.7}   & \textbf{92.8}   & \textbf{94.8}    & \textbf{83.9}   & \textbf{88.6}    & \textbf{74.9}    & \textbf{75.6}  & \textbf{75.7} & \textbf{75.3} & \textbf{70.6} & \textbf{71.1} \\
\hline
\end{tabular}
\begin{tabular}{l||c|cc|cccccc||c}
\hline
      & \multicolumn{1}{c|}{\bf \textsc{Chunk}} & \multicolumn{2}{c|}{\bf \textsc{DP}}   & \multicolumn{6}{c||}{\bf \textsc{SDP}}   & \multirow{2}{*}{\bf \textsc{Avg}}\\
\hhline{~||-|--|------||~}
      & CoNLL 2000         & UAS & LAS  & DM-ID & DM-OOD & PAS-ID & PAS-OOD & PSD-ID & PSD-OOD &           \\
\hline\hline
\bf \textsc{All}    & 96.7    & 96.7     & 95.1 & 94.3  & 90.8   & \textbf{94.6}   & 92.9    & 82.4   & 81.7    & 85.3      \\
\bf \textsc{Random} & 96.7    & 96.8    & 95.2 & 94.4  & 90.8   & \textbf{94.6}   & 93.0    & 82.3   & 81.8    & 85.7      \\
\bf \textsc{ACE}    & \textbf{96.8} & \textbf{96.9}        & \textbf{95.3} & \textbf{94.5}  & \textbf{90.9}   & 94.5   & \textbf{93.1}    & \textbf{82.5}   & \textbf{82.1}    & \textbf{86.2}     \\
\hlineB{4}
\end{tabular}
\caption{Comparison with concatenating all embeddings and random search baselines on 6 tasks.}
\label{tab:baseline}
\end{table*}

\section{Experiments}
We use ISO 639-1 language codes to represent languages in the table\footnote{\url{https://en.wikipedia.org/wiki/List_of_ISO_639-1_codes}}.

\subsection{Datasets and Configurations}
\label{sec:datasets}
To show ACE's effectiveness, we conduct extensive experiments on a variety of structured prediction tasks varying from syntactic tasks to semantic tasks. The tasks are named entity recognition (NER), Part-Of-Speech (POS) tagging, Chunking, Aspect Extraction (AE), Syntactic Dependency Parsing (DP) and Semantic Dependency Parsing (SDP).
The details of the 6 structured prediction tasks in our experiments are shown in below:
\begin{itemize}[leftmargin=*]
    \item {\bf NER}: We use the corpora of 4 languages from the CoNLL 2002 and 2003 shared task \citep{tjong-kim-sang-2002-introduction,tjong-kim-sang-de-meulder-2003-introduction} with standard split.
    \item {\bf POS Tagging}: We use three datasets, Ritter11-T-POS \citep{ritter-etal-2011-named}, ARK-Twitter \citep{gimpel-etal-2011-part,owoputi-etal-2013-improved} and Tweebank-v2 \citep{liu-etal-2018-parsing} datasets (Ritter, ARK and TB-v2 in simplification). We follow the dataset split of \citet{nguyen2020bertweet}.
    \item {\bf Chunking}: We use CoNLL 2000 \citep{tjong-kim-sang-buchholz-2000-introduction} for chunking. Since there is no standard development set for CoNLL 2000 dataset, we split 10\% of the training data as the development set.
    \item {\bf Aspect Extraction}: Aspect extraction is a subtask of aspect-based sentiment analysis \citep{pontiki-etal-2014-semeval,pontiki-etal-2015-semeval,pontiki-etal-2016-semeval}. The datasets are from the laptop and restaurant domain of SemEval 14, restaurant domain of SemEval 15 and restaurant domain of  SemEval 16 shared task (14Lap, 14Res, 15Res and 16Res in short). Additionally, we use another 4 languages in the restaurant domain of SemEval 16 to test our approach in multiple languages. We randomly split 10\% of the training data as the development set following \citet{li-etal-2019-exploiting}.
    \item {\bf Syntactic Dependency Parsing}: We use Penn Tree Bank (PTB) 3.0 with the same dataset pre-processing as \citep{ma-etal-2018-stack}.
    \item {\bf Semantic Dependency Parsing}: We use DM, PAS and PSD datasets for semantic dependency parsing \citep{oepensemeval} for the SemEval 2015 shared task \citep{oepen2015semeval}. The three datasets have the same sentences but with different formalisms. We use the standard split for SDP. In the split, there are in-domain test sets and out-of-domain test sets for each dataset.
\end{itemize}
Among these tasks, NER, POS tagging, chunking and aspect extraction are sequence-structured outputs while dependency parsing and semantic dependency parsing are the graph-structured outputs. POS Tagging, chunking and DP are syntactic structured prediction tasks while NER, AE, SDP are semantic structured prediction tasks.

We train the controller for $30$ steps and save the task model with the highest accuracy on the development set as the final model for testing. Please refer to Appendix \ref{app:detail} for more details of other settings.

\subsection{Embeddings}
\paragraph{Basic Settings:} For the candidates of embeddings on English datasets, we use the language-specific model for ELMo, Flair, base BERT, GloVe word embeddings, fastText word embeddings, non-contextual character embeddings \citep{lample-etal-2016-neural}, multilingual Flair (M-Flair), M-BERT and XLM-R embeddings. The size of the search space in our experiments is $2^{11}{-}1{=}2047$\footnote{Flair embeddings have two models (forward and backward) for each language.}. For language-specific models of other languages, please refer to Appendix \ref{sec:embed} for more details. In AE, there is no available Russian-specific BERT, Flair and ELMo embeddings and there is no available Turkish-specific Flair and ELMo embeddings. We use the corresponding English embeddings instead so that the search spaces of these datasets are almost identical to those of the other datasets. All embeddings are fixed during training except that the character embeddings are trained over the task. The empirical results are reported in Section \ref{sec:exp:base}.

\paragraph{Embedding Fine-tuning:} A usual approach to get better accuracy is fine-tuning transformer-based embeddings. In sequence labeling, most of the work follows the fine-tuning pipeline of BERT that connects the BERT model with a linear layer for word-level classification. 
However, when multiple embeddings are concatenated, fine-tuning a specific group of embeddings becomes difficult because of complicated hyper-parameter settings and massive GPU memory consumption. 
To alleviate this problem, we first fine-tune the transformer-based embeddings over the task and then concatenate these embeddings together with other embeddings in the basic setting to apply ACE. The empirical results are reported in Section \ref{sec:exp:finetune}.


\subsection{Results}
We use the following abbreviations in our experiments: \textbf{UAS}: Unlabeled Attachment Score; \textbf{LAS}: Labeled Attachment Score; \textbf{ID}: In-domain test set; \textbf{OOD}: Out-of-domain test set. We use language codes for languages in NER and AE.

\subsubsection{Comparison With Baselines} \label{sec:exp:base}
To show the effectiveness of our approach, we compare our approach with two strong baselines. For the first one, we let the task model learn by itself the contribution of each embedding candidate that is helpful to the task. We set $\va$ to all-ones (i.e., the concatenation of all the embeddings) and train the task model ({\tt All}). The linear layer weight $\mW$ in Eq. \ref{eq:linear} reflects the contribution of each candidate. For the second one, we use the random search ({\tt Random}), a strong baseline in NAS \citep{li2020random}. For {\tt Random}, we run the same maximum iteration as in ACE. For the experiments, we report the averaged accuracy of 3 runs. Table \ref{tab:baseline} shows that ACE outperforms both baselines in 6 tasks over 23 test sets with only two exceptions. Comparing {\tt Random} with {\tt All}, {\tt Random} outperforms {\tt All} by 0.4 on average and surpasses the accuracy of {\tt All} on 14 out of 23 test sets, which shows that concatenating all embeddings may not be the best solution to most structured prediction tasks. 
In general, searching for the concatenation for the word representation is essential in most cases, and our search design can usually lead to better results compared to both of the baselines.


\begin{table*}[t]
\small
\centering
\begin{tabular}{l|ccccc||l|ccc}
\hlineB{4}
 & \multicolumn{5}{c||}{\bf \textsc{NER}} &       &\multicolumn{3}{c}{\bf \textsc{POS}}\\
 \hhline{~|-----||~|---}
  & de   & de$_\text{06}$ & en   & es   & nl   & & Ritter & ARK  & TB-v2 \\
\hline\hline
\citet{baevski-etal-2019-cloze}      & -    & -    & 93.5 & -    & -    & \citet{owoputi-etal-2013-improved} & 90.4   & 93.2 & 94.6  \\
\citet{strakova-etal-2019-neural}   & 85.1 & -    & 93.4 & 88.8 & 92.7     & \citet{gui-etal-2017-part}          & 90.9   & -    & 92.8  \\
\citet{yu-etal-2020-named} & 86.4 & 90.3 & 93.5 & 90.3 & 93.7 & \citet{gui-etal-2018-transferring} & 91.2   & 92.4 & -     \\
\citet{yamada-etal-2020-luke} & - & - & 94.3 & - & - & \citet{nguyen2020bertweet}         & 90.1   & 94.1 & 95.2  \\
\hline
XLM-R+Fine-tune & 87.7    & 91.4 & 94.1 & 89.3    & 95.3     & XLM-R+Fine-tune    & 92.3   & 93.7 & 95.4 \\
ACE+Fine-tune & \textbf{88.3} & \textbf{91.7} & \textbf{94.6} & \textbf{95.9} & \textbf{95.7} & ACE+Fine-tune  & \textbf{93.4}   & \textbf{94.4} & \textbf{95.8} \\
\hlineB{4}
\end{tabular}
\caption{Comparison with state-of-the-art approaches in NER and POS tagging. $^{\dagger}$: Models are trained on both train and development set.}
\label{tab:ner_pos}
\end{table*}

\begin{table*}[t]
\small
\centering
\setlength\tabcolsep{4pt}
\begin{tabular}{l|c||l|cccccccc}
\hlineB{4}
 & \multicolumn{1}{c||}{\bf \textsc{Chunk}} & &\multicolumn{8}{c}{\bf \textsc{AE}}        \\
\hhline{~|-||~|--------}
 &  CoNLL 2000 & & 14Lap  & 14Res & 15Res & 16Res & es   & nl   & ru   & tr   \\
\hline\hline
\citet{akbik-etal-2018-contextual}  & 96.7    & \citet{xu-etal-2018-double}$^\dagger$ & 84.2   & 84.6 & 72.0  & 75.4  & -    & -    & -    & -    \\
\citet{clark-etal-2018-semi}        & 97.0    & \citet{xu-etal-2019-bert}          & 84.3   & -    & -     & 78.0  & -    & -    & -    & -    \\
\citet{liu-etal-2019-gcdt}              & \textbf{97.3}    & \citet{wang-etal-2020-structure}   & -      & -    & -     & 72.8  & 74.3 & 72.9 & 71.8 & 59.3 \\
\citet{chen-etal-2020-seqvat} & 95.5       & \citet{wei-etal-2020-dont}         & 82.7   & 87.1 & 72.7  & 77.7  & -    & -    & -    & -    \\
\hline
XLM-R+Fine-tune  & 97.0 & XLM-R+Fine-tune & 85.9  & 90.5  & 76.4  & 78.9  & 77.0 & 77.6 & 77.7 & 74.1  \\
ACE+Fine-tune  &  \textbf{97.3}    & ACE+Fine-tune   & \textbf{87.4}   & \textbf{92.0} & \textbf{80.3}  & \textbf{81.3}  & \textbf{79.9} & \textbf{80.5} & \textbf{79.4} & \textbf{81.9} \\
\hlineB{4}
\end{tabular}
\caption{Comparison with state-of-the-art approaches in chunking and aspect extraction. $^\dagger$: We report the results reproduced by \citet{wei-etal-2020-dont}.}
\label{tab:chunk_ae}
\end{table*}

\begin{table*}[t]
\small
\centering
\begin{tabular}{l|cc||l|cccccc}
\hlineB{4}
& \multicolumn{2}{c||}{\bf \textsc{DP}}     &            & \multicolumn{6}{c}{\bf \textsc{SDP}}       \\
\hhline{~|--||~|------}
& \multicolumn{2}{c||}{\bf \textsc{PTB}}     &            & \multicolumn{2}{c}{\bf \textsc{DM}} & \multicolumn{2}{c}{\bf \textsc{PAS}} & \multicolumn{2}{c}{\bf \textsc{PSD}}       \\
     & UAS     & LAS  &            & ID   & OOD & ID  & OOD & ID  & OOD \\
\hline\hline
\citet{zhou-zhao-2019-head}$^\dagger$            & 97.2    & 95.7 &\citet{he2019establishing}$^\ddagger$      &  94.6 & 90.8    & 96.1 & 94.4     & 86.8 & 79.5  \\
\hhline{~|~~||-|------}
\citet{mrini-etal-2020-rethinking}$^\dagger$ & 97.4 & 96.3 &  D \& M \shortcite{dozat-manning-2018-simpler} &93.7 & 88.9    & 93.9 & 90.6     & 81.0 & 79.4  \\
\hhline{-|--||~|~~~~~~}
\citet{li2020global}       & 96.6    & 94.8 &    \citet{wang-etal-2019-second}  & 94.0 & 89.7    & 94.1 & 91.3     & 81.4 & 79.6      \\
 \citet{zhang-etal-2020-efficient}       & 96.1    & 94.5 &      \citet{jia-etal-2020-semi}  &93.6 & 89.1    & - & -    & - & -  \\
  \citet{wang-tu-2020-second} & 96.9    & 95.3& F \& G \shortcite{fernandez-gonzalez-gomez-rodriguez-2020-transition} & 94.4 & 91.0    & 95.1 & 93.4     & 82.6 & 82.0     \\
\hline
XLNET+Fine-tune           & 97.0        & 95.6   &XLNet+Fine-tune    & 94.2       & 90.6      & 94.8       & 93.4       & 82.7       & 81.8\\
ACE+Fine-tune  & \textbf{97.2}    & \textbf{95.8} & ACE+Fine-tune  & \textbf{95.6} & \textbf{92.6}    & \textbf{95.8} & \textbf{94.6}     & \textbf{83.8} & \textbf{83.4}    \\
\hlineB{4}
\end{tabular}
\caption{Comparison with state-of-the-art approaches in DP and SDP. $^\dagger$: For reference, they additionally used constituency dependencies in training. We also find that the PTB dataset used by \citet{mrini-etal-2020-rethinking} is not identical to the dataset in previous work such as \citet{zhang-etal-2020-efficient} and \citet{wang-tu-2020-second}. $^\ddagger$: For reference, we confirmed with the authors of \citet{he2019establishing} that they used a different data pre-processing script with previous work.}.
\label{tab:dp_sdp}
\end{table*}

\subsubsection{Comparison With State-of-the-Art approaches} \label{sec:exp:finetune}
As we have shown, ACE has an advantage in searching for better embedding concatenations. We further show that ACE is competitive or even stronger than state-of-the-art approaches. We additionally use XLNet \citep{yang2019xlnet} and RoBERTa as the candidates of ACE.  In some tasks, we have several additional settings to better compare with previous work. In NER, we also conduct a comparison on the revised version of German datasets in the CoNLL 2006 shared task \citep{buchholz-marsi-2006-conll}. Recent work such as \citet{yu-etal-2020-named} and \citet{yamada-etal-2020-luke} utilizes document contexts in the datasets. We follow their work and extract document embeddings for the transformer-based embeddings. Specifically, we follow the fine-tune process of \citet{yamada-etal-2020-luke} to fine-tune the transformer-based embeddings over the document except for BERT and M-BERT embeddings. For BERT and M-BERT, we follow the document extraction process of \citet{yu-etal-2020-named} because we find that the model with such document embeddings is significantly stronger than the model trained with the fine-tuning process of \citet{yamada-etal-2020-luke}. In SDP, the state-of-the-art approaches used POS tags and lemmas as additional word features to the network. We add these two features to the embedding candidates and train the embeddings together with the task. We use the fine-tuned transformer-based embeddings on each task instead of the pretrained version of these embeddings as the candidates.\footnote{Please refer to Appendix for more details about the embeddings.}

We additionally compare with fine-tuned XLM-R model for NER, POS tagging, chunking and AE, and compare with fine-tuned XLNet model for DP and SDP, which are strong fine-tuned models in most of the experiments. Results are shown in Table \ref{tab:ner_pos}, \ref{tab:chunk_ae}, \ref{tab:dp_sdp}.
Results show that ACE with fine-tuned embeddings achieves state-of-the-art performance in all test sets, which shows that finding a good embedding concatenation helps structured prediction tasks. We also find that ACE is stronger than the fine-tuned models, which shows the effectiveness of concatenating the fine-tuned embeddings\footnote{We compare ACE with other fine-tuned embeddings in Appendix.}.

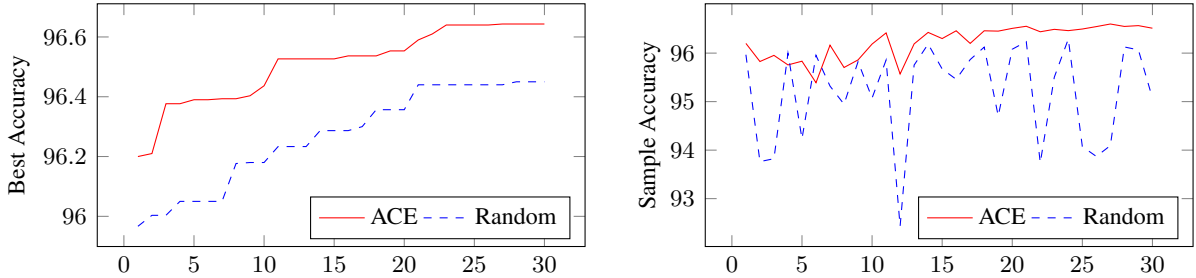
\begin{figure*}[t!]
\begin{minipage}{1.0\linewidth}
\centering
\begin{tikzpicture}
    \begin{axis}[
        xshift=-6cm,
        name=ner,
        width=0.5\textwidth,
        height=0.3\textwidth,
        ylabel=Best Accuracy,
        legend columns=2, 
        legend pos=south east,
        legend style={font=\small},
        tick label style={font=\small},
        ylabel style={font=\small,yshift=-0.2cm},
        ]
        \addplot[red] table[x=episode,y=ACE] {max_dev.dat};
        \addplot[blue,dashed] table[x=episode,y=Random] {max_dev.dat};
        \legend{ACE, Random}
    \end{axis}
    \begin{axis}[
        at={(ner.south west)},
        xshift=8.0cm,
        width=0.5\textwidth,
        height=0.3\textwidth,
        ylabel=Sample Accuracy,
        legend columns=2, 
        legend pos=south east,
        legend style={font=\small},
        tick label style={font=\small},
        ytick={91,92,93,94,95,96},
        ylabel style={font=\small,yshift=-0.5cm},
        ]
        \addplot[red] table[x=episode,y=ACE] {avg_dev.dat};
        \addplot[blue,dashed] table[x=episode,y=Random] {avg_dev.dat};
        \legend{ACE, Random}
    \end{axis}
\end{tikzpicture}
\caption{Comparing the efficiency of random search ({\tt Random}) and ACE. The x-axis is the number of time steps. The left y-axis is the averaged best validation accuracy on CoNLL English NER dataset. The right y-axis is the averaged validation accuracy of the current selection.}
\label{fig:dev_curve}
\end{minipage}
\end{figure*}


\section{Analysis}

\subsection{Efficiency of Search Methods}
To show how efficient our approach is compared with the random search algorithm, we compare the algorithm in two aspects on CoNLL English NER dataset. The first aspect is the best development accuracy during training. The left part of Figure \ref{fig:dev_curve} shows that ACE is consistently stronger than the random search algorithm in this task. The second aspect is the searched concatenation at each time step. The right part of Figure \ref{fig:dev_curve} shows that the accuracy of ACE gradually increases and gets stable when more concatenations are sampled.

\subsection{Ablation Study on Reward Function Design}
\label{sec:ablation}
To show the effectiveness of the designed reward function, we compare our reward function (Eq. \ref{eq:reward}) with the reward function without discount factor (Eq. \ref{eq:reward_a}) and the traditional reward function (reward term in Eq. \ref{eq:rl}). We sample 2000 training sentences on CoNLL English NER dataset for faster training and train the controller for $50$ steps. Table \ref{tab:ablation} shows that both the discount factor and the binary vector $|\va^t-\va^{i}|$ for the task are helpful in both development and test datasets. 
\begin{table}[h!]
\small
\centering
\begin{tabular}{l||cc}
\hlineB{4}
 & \bf \textsc{Dev} &   \bf \textsc{Test}\\
\hline
ACE & \textbf{93.18} & \textbf{90.00}\\
No discount (Eq. \ref{eq:reward_a}) &  92.98 & 89.90\\
Simple (Eq. \ref{eq:rl}) & 92.89 & 89.82 \\
\hlineB{4}
\end{tabular}
\caption{Comparison of reward functions.}
\label{tab:ablation}
\end{table}

\begin{table}[!ht]
\small
\centering
\setlength\tabcolsep{1.2pt}
\begin{tabular}{l||cccccccc}
\hlineB{4}
             & \multirow{2}{*}{\bf \textsc{NER}}  & \multirow{2}{*}{\bf \textsc{POS}}  &  \multirow{2}{*}{\bf \textsc{AE}}   & \multirow{2}{*}{\bf \textsc{CHK}} &\multicolumn{2}{c}{\bf \textsc{DP}} & \multicolumn{2}{c}{\bf \textsc{SDP}}  \\
&  &  & &   & UAS & LAS & ID  & OOD \\
\hline
\hline
All          & 92.4 & 90.6 & 73.2 & 96.7  & 96.7   & 95.1   & 94.3 & 90.8    \\
Random          & 92.6 & 91.3  & 74.7 & 96.7 & 96.8   & 95.2   & 94.4 & 90.8    \\
ACE          & \textbf{93.0} & \textbf{91.7} & \textbf{75.6}& \textbf{96.8}   & \textbf{96.9}   & \textbf{95.3}   & \textbf{94.5} & \textbf{90.9}   \\
All+Weight & 92.7  & 90.4 & 73.7 & 96.7 & 96.7 & 95.1 & 94.3 & 90.7 \\ 
Ensemble & 92.2 & 90.6  & 68.1 & 96.5 & 96.1   &94.3   & 94.1 & 90.3    \\ 
Ensemble$_{\text{dev}}$ & 92.2 & 90.8   & 70.2 & 96.7 & 96.8   &95.2   & 94.3 & 90.7    \\
\hline
Ensemble$_{\text{test}}$ & 92.7 & 91.4  & 73.9 & 96.7 & 96.8   &95.2   & 94.4 & 90.8    \\
\hlineB{4}
\end{tabular}
\caption{A comparison among \texttt{All}, \texttt{Random}, ACE, \texttt{All+Weight} and \texttt{Ensemble}. \textbf{\sc CHK}: chunking.}
\label{tab:ensemble}
\end{table}

\subsection{Comparison with Embedding Weighting \& Ensemble Approaches}
\label{sec:ensemble}
We compare ACE with two more approaches to further show the effectiveness of ACE. One is a variant of \texttt{All}, which uses a weighting parameter $\vb=[b_1, \cdots, b_l, \cdots, b_L]$ passing through a sigmoid function to weight each embedding candidate. Such an approach can explicitly learn the weight of each embedding in training instead of a binary mask. We call this approach \texttt{All+Weight}. Another one is model ensemble, which trains the task model with each embedding candidate individually and uses the trained models to make joint prediction on the test set. We use voting for ensemble as it is simple and fast. For sequence labeling tasks, the models vote for the predicted label at each position. For DP, the models vote for the tree of each sentence. For SDP, the models vote for each potential labeled arc. We use the confidence of model predictions to break ties if there are more than one agreement with the same counts. We call this approach \texttt{Ensemble}. One of the benefits of voting is that it combines the predictions of the task models efficiently without any training process. We can search all possible $2^L{-}1$ model ensembles in a short period of time through caching the outputs of the models. Therefore, we search for the best ensemble of models on the development set and then evaluate the best ensemble on the test set (\texttt{Ensemble$_{\texttt{dev}}$}). Moreover, we additionally search for the best ensemble on the test set for reference (\texttt{Ensemble$_{\texttt{test}}$}), which is the upper bound of the approach. We use the same setting as in Section \ref{sec:exp:base} and select one of the datasets from each task. For NER, POS tagging, AE, and SDP, we use CoNLL 2003 English, Ritter, 16Res, and DM datasets, respectively. The results are shown in Table \ref{tab:ensemble}. Empirical results show that ACE outperforms all the settings of these approaches and even \texttt{Ensemble$_\texttt{test}$}, which shows the effectiveness of ACE and the limitation of ensemble models. \texttt{All}, \texttt{All+Weight} and \texttt{Ensemble$_\texttt{dev}$} are competitive in most of the cases and there is no clear winner of these approaches on all the datasets. These results show the strength of embedding concatenation. Concatenating the embeddings incorporates information from all the embeddings and forms stronger word representations for the task model, while in model ensemble, it is difficult for the individual task models to affect each other.

\section{Discussion: Practical Usability of ACE}
Concatenating multiple embeddings is a commonly used approach to improve accuracy of structured prediction. However, such approaches can be computationally costly as multiple language models are used as input. ACE is more practical than concatenating all embeddings as it can remove those embeddings that are not very useful in the concatenation. Moreover, ACE models can be used to guide the training of weaker models through techniques such as knowledge distillation in structured prediction \cite{kim-rush-2016-sequence,kuncoro-etal-2016-distilling,wang-etal-2020-structure,wang2020structural}, leading to models that are both stronger and faster. 

\section{Conclusion}
In this paper, we propose Automated Concatenation of Embeddings, which automatically searches for better embedding concatenation for structured prediction tasks. We design a simple search space and use the reinforcement learning with a novel reward function to efficiently guide the controller to search for better embedding concatenations. We take the change of embedding concatenations into the reward function design and show that our new reward function is stronger than the simpler ones. Results show that ACE outperforms strong baselines. Together with fine-tuned embeddings, ACE achieves state-of-the-art performance in 6 tasks over 21 datasets.

\subsubsection*{Acknowledgments}
This work was supported by the National Natural Science Foundation of China (61976139) and by Alibaba Group through Alibaba Innovative Research Program. We thank Chengyue Jiang for his comments and suggestions on writing. 

\bibliographystyle{acl_natbib}
\bibliography{anthology,acl2021}

\begin{thebibliography}{101}
\expandafter\ifx\csname natexlab\endcsname\relax\def\natexlab#1{#1}\fi

\bibitem[{Akbik et~al.(2019)Akbik, Bergmann, and
  Vollgraf}]{akbik-etal-2019-pooled}
Alan Akbik, Tanja Bergmann, and Roland Vollgraf. 2019.
\newblock \href {https://doi.org/10.18653/v1/N19-1078} {Pooled contextualized
  embeddings for named entity recognition}.
\newblock In \emph{Proceedings of the 2019 Conference of the North {A}merican
  Chapter of the Association for Computational Linguistics: Human Language
  Technologies, Volume 1 (Long and Short Papers)}, pages 724--728, Minneapolis,
  Minnesota. Association for Computational Linguistics.

\bibitem[{Akbik et~al.(2018)Akbik, Blythe, and
  Vollgraf}]{akbik-etal-2018-contextual}
Alan Akbik, Duncan Blythe, and Roland Vollgraf. 2018.
\newblock \href {https://www.aclweb.org/anthology/C18-1139} {Contextual string
  embeddings for sequence labeling}.
\newblock In \emph{Proceedings of the 27th International Conference on
  Computational Linguistics}, pages 1638--1649, Santa Fe, New Mexico, USA.
  Association for Computational Linguistics.

\bibitem[{Angeline et~al.(1994)Angeline, Saunders, and
  Pollack}]{angeline1994evolutionary}
Peter~J Angeline, Gregory~M Saunders, and Jordan~B Pollack. 1994.
\newblock An evolutionary algorithm that constructs recurrent neural networks.
\newblock \emph{IEEE transactions on Neural Networks}, 5(1):54--65.

\bibitem[{Baevski et~al.(2019)Baevski, Edunov, Liu, Zettlemoyer, and
  Auli}]{baevski-etal-2019-cloze}
Alexei Baevski, Sergey Edunov, Yinhan Liu, Luke Zettlemoyer, and Michael Auli.
  2019.
\newblock \href {https://doi.org/10.18653/v1/D19-1539} {Cloze-driven
  pretraining of self-attention networks}.
\newblock In \emph{Proceedings of the 2019 Conference on Empirical Methods in
  Natural Language Processing and the 9th International Joint Conference on
  Natural Language Processing (EMNLP-IJCNLP)}, pages 5360--5369, Hong Kong,
  China. Association for Computational Linguistics.

\bibitem[{Baker et~al.(2017)Baker, Gupta, Naik, and
  Raskar}]{baker2016designing}
Bowen Baker, Otkrist Gupta, Nikhil Naik, and Ramesh Raskar. 2017.
\newblock Designing neural network architectures using reinforcement learning.
\newblock In \emph{International Conference on Learning Representations}.

\bibitem[{Bojanowski et~al.(2017)Bojanowski, Grave, Joulin, and
  Mikolov}]{bojanowski2017enriching}
Piotr Bojanowski, Edouard Grave, Armand Joulin, and Tomas Mikolov. 2017.
\newblock Enriching word vectors with subword information.
\newblock \emph{Transactions of the Association for Computational Linguistics},
  5:135--146.

\bibitem[{Buchholz and Marsi(2006)}]{buchholz-marsi-2006-conll}
Sabine Buchholz and Erwin Marsi. 2006.
\newblock \href {https://www.aclweb.org/anthology/W06-2920} {{C}o{NLL}-{X}
  shared task on multilingual dependency parsing}.
\newblock In \emph{Proceedings of the Tenth Conference on Computational Natural
  Language Learning ({C}o{NLL}-X)}, pages 149--164, New York City. Association
  for Computational Linguistics.

\bibitem[{Chen et~al.(2020)Chen, Ruan, Liu, and Lu}]{chen-etal-2020-seqvat}
Luoxin Chen, Weitong Ruan, Xinyue Liu, and Jianhua Lu. 2020.
\newblock \href {https://doi.org/10.18653/v1/2020.acl-main.777} {{S}eq{VAT}:
  Virtual adversarial training for semi-supervised sequence labeling}.
\newblock In \emph{Proceedings of the 58th Annual Meeting of the Association
  for Computational Linguistics}, pages 8801--8811, Online. Association for
  Computational Linguistics.

\bibitem[{Clark et~al.(2018)Clark, Luong, Manning, and
  Le}]{clark-etal-2018-semi}
Kevin Clark, Minh-Thang Luong, Christopher~D. Manning, and Quoc Le. 2018.
\newblock \href {https://doi.org/10.18653/v1/D18-1217} {Semi-supervised
  sequence modeling with cross-view training}.
\newblock In \emph{Proceedings of the 2018 Conference on Empirical Methods in
  Natural Language Processing}, pages 1914--1925, Brussels, Belgium.
  Association for Computational Linguistics.

\bibitem[{Conneau et~al.(2020)Conneau, Khandelwal, Goyal, Chaudhary, Wenzek,
  Guzm{\'a}n, Grave, Ott, Zettlemoyer, and
  Stoyanov}]{conneau-etal-2020-unsupervised}
Alexis Conneau, Kartikay Khandelwal, Naman Goyal, Vishrav Chaudhary, Guillaume
  Wenzek, Francisco Guzm{\'a}n, Edouard Grave, Myle Ott, Luke Zettlemoyer, and
  Veselin Stoyanov. 2020.
\newblock \href {https://doi.org/10.18653/v1/2020.acl-main.747} {Unsupervised
  cross-lingual representation learning at scale}.
\newblock In \emph{Proceedings of the 58th Annual Meeting of the Association
  for Computational Linguistics}, pages 8440--8451, Online. Association for
  Computational Linguistics.

\bibitem[{DeRose(1988)}]{derose-1988-grammatical}
Steven~J. DeRose. 1988.
\newblock \href {https://www.aclweb.org/anthology/J88-1003} {Grammatical
  category disambiguation by statistical optimization}.
\newblock \emph{Computational Linguistics}, 14(1):31--39.

\bibitem[{Devlin et~al.(2019)Devlin, Chang, Lee, and
  Toutanova}]{devlin-etal-2019-bert}
Jacob Devlin, Ming-Wei Chang, Kenton Lee, and Kristina Toutanova. 2019.
\newblock \href {https://doi.org/10.18653/v1/N19-1423} {{BERT}: Pre-training of
  deep bidirectional transformers for language understanding}.
\newblock In \emph{Proceedings of the 2019 Conference of the North {A}merican
  Chapter of the Association for Computational Linguistics: Human Language
  Technologies, Volume 1 (Long and Short Papers)}, pages 4171--4186,
  Minneapolis, Minnesota. Association for Computational Linguistics.

\bibitem[{Dozat and Manning(2017)}]{dozat2016deep}
Timothy Dozat and Christopher~D Manning. 2017.
\newblock Deep biaffine attention for neural dependency parsing.
\newblock In \emph{International Conference on Learning Representations}.

\bibitem[{Dozat and Manning(2018)}]{dozat-manning-2018-simpler}
Timothy Dozat and Christopher~D. Manning. 2018.
\newblock \href {https://doi.org/10.18653/v1/P18-2077} {Simpler but more
  accurate semantic dependency parsing}.
\newblock In \emph{Proceedings of the 56th Annual Meeting of the Association
  for Computational Linguistics (Volume 2: Short Papers)}, pages 484--490,
  Melbourne, Australia. Association for Computational Linguistics.

\bibitem[{Elsken et~al.(2018)Elsken, Metzen, and Hutter}]{elsken2017simple}
Thomas Elsken, Jan-Hendrik Metzen, and Frank Hutter. 2018.
\newblock Simple and efficient architecture search for convolutional neural
  networks.
\newblock In \emph{International Conference on Learning Representations
  workshop}.

\bibitem[{Elsken et~al.(2019)Elsken, Metzen, and Hutter}]{elsken2019neural}
Thomas Elsken, Jan~Hendrik Metzen, and Frank Hutter. 2019.
\newblock Neural architecture search: A survey.
\newblock \emph{Journal of Machine Learning Research}, 20:1--21.

\bibitem[{Fern{\'a}ndez-Gonz{\'a}lez and
  G{\'o}mez-Rodr{\'\i}guez(2020)}]{fernandez-gonzalez-gomez-rodriguez-2020-transition}
Daniel Fern{\'a}ndez-Gonz{\'a}lez and Carlos G{\'o}mez-Rodr{\'\i}guez. 2020.
\newblock \href {https://doi.org/10.18653/v1/2020.acl-main.629}
  {Transition-based semantic dependency parsing with pointer networks}.
\newblock In \emph{Proceedings of the 58th Annual Meeting of the Association
  for Computational Linguistics}, pages 7035--7046, Online. Association for
  Computational Linguistics.

\bibitem[{Floreano et~al.(2008)Floreano, D{\"u}rr, and
  Mattiussi}]{floreano2008neuroevolution}
Dario Floreano, Peter D{\"u}rr, and Claudio Mattiussi. 2008.
\newblock Neuroevolution: from architectures to learning.
\newblock \emph{Evolutionary intelligence}, 1(1):47--62.

\bibitem[{Ghiasi et~al.(2019)Ghiasi, Lin, and Le}]{ghiasi2019fpn}
Golnaz Ghiasi, Tsung-Yi Lin, and Quoc~V Le. 2019.
\newblock Nas-fpn: Learning scalable feature pyramid architecture for object
  detection.
\newblock In \emph{Proceedings of the IEEE conference on computer vision and
  pattern recognition}, pages 7036--7045.

\bibitem[{Gimpel et~al.(2011)Gimpel, Schneider, O{'}Connor, Das, Mills,
  Eisenstein, Heilman, Yogatama, Flanigan, and Smith}]{gimpel-etal-2011-part}
Kevin Gimpel, Nathan Schneider, Brendan O{'}Connor, Dipanjan Das, Daniel Mills,
  Jacob Eisenstein, Michael Heilman, Dani Yogatama, Jeffrey Flanigan, and
  Noah~A. Smith. 2011.
\newblock \href {https://www.aclweb.org/anthology/P11-2008} {Part-of-speech
  tagging for {T}witter: Annotation, features, and experiments}.
\newblock In \emph{Proceedings of the 49th Annual Meeting of the Association
  for Computational Linguistics: Human Language Technologies}, pages 42--47,
  Portland, Oregon, USA. Association for Computational Linguistics.

\bibitem[{Goldberg and Deb(1991)}]{goldberg1991comparative}
David~E Goldberg and Kalyanmoy Deb. 1991.
\newblock A comparative analysis of selection schemes used in genetic
  algorithms.
\newblock In \emph{Foundations of genetic algorithms}, volume~1, pages 69--93.
  Elsevier.

\bibitem[{Gui et~al.(2018)Gui, Zhang, Gong, Peng, Liang, Ding, and
  Huang}]{gui-etal-2018-transferring}
Tao Gui, Qi~Zhang, Jingjing Gong, Minlong Peng, Di~Liang, Keyu Ding, and
  Xuanjing Huang. 2018.
\newblock \href {https://doi.org/10.18653/v1/D18-1275} {Transferring from
  formal newswire domain with hypernet for {T}witter {POS} tagging}.
\newblock In \emph{Proceedings of the 2018 Conference on Empirical Methods in
  Natural Language Processing}, pages 2540--2549, Brussels, Belgium.
  Association for Computational Linguistics.

\bibitem[{Gui et~al.(2017)Gui, Zhang, Huang, Peng, and
  Huang}]{gui-etal-2017-part}
Tao Gui, Qi~Zhang, Haoran Huang, Minlong Peng, and Xuanjing Huang. 2017.
\newblock \href {https://doi.org/10.18653/v1/D17-1256} {Part-of-speech tagging
  for {T}witter with adversarial neural networks}.
\newblock In \emph{Proceedings of the 2017 Conference on Empirical Methods in
  Natural Language Processing}, pages 2411--2420, Copenhagen, Denmark.
  Association for Computational Linguistics.

\bibitem[{He and Choi(2020)}]{he2019establishing}
Han He and Jinho Choi. 2020.
\newblock Establishing strong baselines for the new decade: Sequence tagging,
  syntactic and semantic parsing with bert.
\newblock In \emph{The Thirty-Third International Flairs Conference}.

\bibitem[{He et~al.(2016)He, Zhang, Ren, and Sun}]{he2016deep}
Kaiming He, Xiangyu Zhang, Shaoqing Ren, and Jian Sun. 2016.
\newblock Deep residual learning for image recognition.
\newblock In \emph{Proceedings of the IEEE conference on computer vision and
  pattern recognition}, pages 770--778.

\bibitem[{Hu and Liu(2004)}]{10.1145/1014052.1014073}
Minqing Hu and Bing Liu. 2004.
\newblock \href {https://doi.org/10.1145/1014052.1014073} {Mining and
  summarizing customer reviews}.
\newblock In \emph{Proceedings of the Tenth ACM SIGKDD International Conference
  on Knowledge Discovery and Data Mining}, KDD '04, page 168–177, New York,
  NY, USA. Association for Computing Machinery.

\bibitem[{Huang et~al.(2017)Huang, Liu, Van Der~Maaten, and
  Weinberger}]{huang2017densely}
Gao Huang, Zhuang Liu, Laurens Van Der~Maaten, and Kilian~Q Weinberger. 2017.
\newblock Densely connected convolutional networks.
\newblock In \emph{Proceedings of the IEEE conference on computer vision and
  pattern recognition}, pages 4700--4708.

\bibitem[{Jia et~al.(2020)Jia, Ma, Cai, and Tu}]{jia-etal-2020-semi}
Zixia Jia, Youmi Ma, Jiong Cai, and Kewei Tu. 2020.
\newblock \href {https://doi.org/10.18653/v1/2020.acl-main.607}
  {Semi-supervised semantic dependency parsing using {CRF} autoencoders}.
\newblock In \emph{Proceedings of the 58th Annual Meeting of the Association
  for Computational Linguistics}, pages 6795--6805, Online. Association for
  Computational Linguistics.

\bibitem[{Jozefowicz et~al.(2015)Jozefowicz, Zaremba, and
  Sutskever}]{jozefowicz2015empirical}
Rafal Jozefowicz, Wojciech Zaremba, and Ilya Sutskever. 2015.
\newblock An empirical exploration of recurrent network architectures.
\newblock In \emph{International conference on machine learning}, pages
  2342--2350.

\bibitem[{Kim and Rush(2016)}]{kim-rush-2016-sequence}
Yoon Kim and Alexander~M. Rush. 2016.
\newblock \href {https://doi.org/10.18653/v1/D16-1139} {Sequence-level
  knowledge distillation}.
\newblock In \emph{Proceedings of the 2016 Conference on Empirical Methods in
  Natural Language Processing}, pages 1317--1327, Austin, Texas. Association
  for Computational Linguistics.

\bibitem[{Kingma and Ba(2015)}]{kingma2014adam}
Diederik~P Kingma and Jimmy Ba. 2015.
\newblock Adam: A method for stochastic optimization.
\newblock In \emph{International Conference on Learning Representations}.

\bibitem[{Kondratyuk and Straka(2019)}]{kondratyuk-straka-2019-75}
Dan Kondratyuk and Milan Straka. 2019.
\newblock \href {https://doi.org/10.18653/v1/D19-1279} {75 languages, 1 model:
  Parsing {U}niversal {D}ependencies universally}.
\newblock In \emph{Proceedings of the 2019 Conference on Empirical Methods in
  Natural Language Processing and the 9th International Joint Conference on
  Natural Language Processing (EMNLP-IJCNLP)}, pages 2779--2795, Hong Kong,
  China. Association for Computational Linguistics.

\bibitem[{Kuncoro et~al.(2016)Kuncoro, Ballesteros, Kong, Dyer, and
  Smith}]{kuncoro-etal-2016-distilling}
Adhiguna Kuncoro, Miguel Ballesteros, Lingpeng Kong, Chris Dyer, and Noah~A.
  Smith. 2016.
\newblock \href {https://doi.org/10.18653/v1/D16-1180} {Distilling an ensemble
  of greedy dependency parsers into one {MST} parser}.
\newblock In \emph{Proceedings of the 2016 Conference on Empirical Methods in
  Natural Language Processing}, pages 1744--1753, Austin, Texas. Association
  for Computational Linguistics.

\bibitem[{Lample et~al.(2016)Lample, Ballesteros, Subramanian, Kawakami, and
  Dyer}]{lample-etal-2016-neural}
Guillaume Lample, Miguel Ballesteros, Sandeep Subramanian, Kazuya Kawakami, and
  Chris Dyer. 2016.
\newblock \href {https://doi.org/10.18653/v1/N16-1030} {Neural architectures
  for named entity recognition}.
\newblock In \emph{Proceedings of the 2016 Conference of the North {A}merican
  Chapter of the Association for Computational Linguistics: Human Language
  Technologies}, pages 260--270, San Diego, California. Association for
  Computational Linguistics.

\bibitem[{Li and Talwalkar(2020)}]{li2020random}
Liam Li and Ameet Talwalkar. 2020.
\newblock Random search and reproducibility for neural architecture search.
\newblock In \emph{Uncertainty in Artificial Intelligence}, pages 367--377.
  PMLR.

\bibitem[{Li et~al.(2019)Li, Bing, Zhang, and Lam}]{li-etal-2019-exploiting}
Xin Li, Lidong Bing, Wenxuan Zhang, and Wai Lam. 2019.
\newblock \href {https://doi.org/10.18653/v1/D19-5505} {Exploiting {BERT} for
  end-to-end aspect-based sentiment analysis}.
\newblock In \emph{Proceedings of the 5th Workshop on Noisy User-generated Text
  (W-NUT 2019)}, pages 34--41, Hong Kong, China. Association for Computational
  Linguistics.

\bibitem[{Li et~al.(2020)Li, Zhao, and Parnow}]{li2020global}
Zuchao Li, Hai Zhao, and Kevin Parnow. 2020.
\newblock Global greedy dependency parsing.
\newblock In \emph{Proceedings of the AAAI Conference on Artificial
  Intelligence}, volume~34, pages 8319--8326.

\bibitem[{Liu et~al.(2019{\natexlab{a}})Liu, Chen, Schroff, Adam, Hua, Yuille,
  and Fei-Fei}]{liu2019auto}
Chenxi Liu, Liang-Chieh Chen, Florian Schroff, Hartwig Adam, Wei Hua, Alan~L
  Yuille, and Li~Fei-Fei. 2019{\natexlab{a}}.
\newblock Auto-deeplab: Hierarchical neural architecture search for semantic
  image segmentation.
\newblock In \emph{Proceedings of the IEEE conference on computer vision and
  pattern recognition}, pages 82--92.

\bibitem[{Liu et~al.(2018{\natexlab{a}})Liu, Simonyan, Vinyals, Fernando, and
  Kavukcuoglu}]{liu2018hierarchical}
Hanxiao Liu, Karen Simonyan, Oriol Vinyals, Chrisantha Fernando, and Koray
  Kavukcuoglu. 2018{\natexlab{a}}.
\newblock Hierarchical representations for efficient architecture search.
\newblock In \emph{International Conference on Learning Representations}.

\bibitem[{Liu et~al.(2018{\natexlab{b}})Liu, Zhu, Che, Qin, Schneider, and
  Smith}]{liu-etal-2018-parsing}
Yijia Liu, Yi~Zhu, Wanxiang Che, Bing Qin, Nathan Schneider, and Noah~A. Smith.
  2018{\natexlab{b}}.
\newblock \href {https://doi.org/10.18653/v1/N18-1088} {Parsing tweets into
  {U}niversal {D}ependencies}.
\newblock In \emph{Proceedings of the 2018 Conference of the North {A}merican
  Chapter of the Association for Computational Linguistics: Human Language
  Technologies, Volume 1 (Long Papers)}, pages 965--975, New Orleans,
  Louisiana. Association for Computational Linguistics.

\bibitem[{Liu et~al.(2019{\natexlab{b}})Liu, Meng, Zhang, Xu, Chen, and
  Zhou}]{liu-etal-2019-gcdt}
Yijin Liu, Fandong Meng, Jinchao Zhang, Jinan Xu, Yufeng Chen, and Jie Zhou.
  2019{\natexlab{b}}.
\newblock \href {https://doi.org/10.18653/v1/P19-1233} {{GCDT}: A global
  context enhanced deep transition architecture for sequence labeling}.
\newblock In \emph{Proceedings of the 57th Annual Meeting of the Association
  for Computational Linguistics}, pages 2431--2441, Florence, Italy.
  Association for Computational Linguistics.

\bibitem[{Liu et~al.(2019{\natexlab{c}})Liu, Ott, Goyal, Du, Joshi, Chen, Levy,
  Lewis, Zettlemoyer, and Stoyanov}]{liu2019roberta}
Yinhan Liu, Myle Ott, Naman Goyal, Jingfei Du, Mandar Joshi, Danqi Chen, Omer
  Levy, Mike Lewis, Luke Zettlemoyer, and Veselin Stoyanov. 2019{\natexlab{c}}.
\newblock Roberta: A robustly optimized bert pretraining approach.
\newblock \emph{arXiv preprint arXiv:1907.11692}.

\bibitem[{Loshchilov and Hutter(2018)}]{loshchilov2018decoupled}
Ilya Loshchilov and Frank Hutter. 2018.
\newblock Decoupled weight decay regularization.
\newblock In \emph{International Conference on Learning Representations}.

\bibitem[{Luoma and Pyysalo(2020)}]{luoma-pyysalo-2020-exploring}
Jouni Luoma and Sampo Pyysalo. 2020.
\newblock \href {https://www.aclweb.org/anthology/2020.coling-main.78}
  {Exploring cross-sentence contexts for named entity recognition with {BERT}}.
\newblock In \emph{Proceedings of the 28th International Conference on
  Computational Linguistics}, pages 904--914, Barcelona, Spain (Online).
  International Committee on Computational Linguistics.

\bibitem[{Ma and Hovy(2016)}]{ma-hovy-2016-end}
Xuezhe Ma and Eduard Hovy. 2016.
\newblock \href {https://doi.org/10.18653/v1/P16-1101} {End-to-end sequence
  labeling via bi-directional {LSTM}-{CNN}s-{CRF}}.
\newblock In \emph{Proceedings of the 54th Annual Meeting of the Association
  for Computational Linguistics (Volume 1: Long Papers)}, pages 1064--1074,
  Berlin, Germany. Association for Computational Linguistics.

\bibitem[{Ma et~al.(2018)Ma, Hu, Liu, Peng, Neubig, and
  Hovy}]{ma-etal-2018-stack}
Xuezhe Ma, Zecong Hu, Jingzhou Liu, Nanyun Peng, Graham Neubig, and Eduard
  Hovy. 2018.
\newblock \href {https://doi.org/10.18653/v1/P18-1130} {Stack-pointer networks
  for dependency parsing}.
\newblock In \emph{Proceedings of the 56th Annual Meeting of the Association
  for Computational Linguistics (Volume 1: Long Papers)}, pages 1403--1414,
  Melbourne, Australia. Association for Computational Linguistics.

\bibitem[{McDonald et~al.(2005)McDonald, Pereira, Ribarov, and
  Haji{\v{c}}}]{mcdonald-etal-2005-non}
Ryan McDonald, Fernando Pereira, Kiril Ribarov, and Jan Haji{\v{c}}. 2005.
\newblock \href {https://www.aclweb.org/anthology/H05-1066} {Non-projective
  dependency parsing using spanning tree algorithms}.
\newblock In \emph{Proceedings of Human Language Technology Conference and
  Conference on Empirical Methods in Natural Language Processing}, pages
  523--530, Vancouver, British Columbia, Canada. Association for Computational
  Linguistics.

\bibitem[{Mikolov et~al.(2013)Mikolov, Sutskever, Chen, Corrado, and
  Dean}]{mikolov2013distributed}
Tomas Mikolov, Ilya Sutskever, Kai Chen, Greg~S Corrado, and Jeff Dean. 2013.
\newblock Distributed representations of words and phrases and their
  compositionality.
\newblock In \emph{Advances in neural information processing systems}, pages
  3111--3119.

\bibitem[{Miller et~al.(1989)Miller, Todd, and Hegde}]{geoffrey1989designing}
Geoffrey Miller, Peter Todd, and Shailesh Hegde. 1989.
\newblock Designing neural networks using genetic algorithms.
\newblock In \emph{3rd International Conference on Genetic Algorithms}, pages
  379--384.

\bibitem[{Mrini et~al.(2020)Mrini, Dernoncourt, Tran, Bui, Chang, and
  Nakashole}]{mrini-etal-2020-rethinking}
Khalil Mrini, Franck Dernoncourt, Quan~Hung Tran, Trung Bui, Walter Chang, and
  Ndapa Nakashole. 2020.
\newblock \href {https://doi.org/10.18653/v1/2020.findings-emnlp.65}
  {Rethinking self-attention: Towards interpretability in neural parsing}.
\newblock In \emph{Findings of the Association for Computational Linguistics:
  EMNLP 2020}, pages 731--742, Online. Association for Computational
  Linguistics.

\bibitem[{Nguyen et~al.(2020)Nguyen, Vu, and Nguyen}]{nguyen2020bertweet}
Dat~Quoc Nguyen, Thanh Vu, and Anh~Tuan Nguyen. 2020.
\newblock {BERTweet: A pre-trained language model for English Tweets}.
\newblock In \emph{Proceedings of the 2020 Conference on Empirical Methods in
  Natural Language Processing: System Demonstrations}.

\bibitem[{Oepen et~al.(2015)Oepen, Kuhlmann, Miyao, Zeman, Cinkov{\'a},
  Flickinger, Hajic, and Uresova}]{oepen2015semeval}
Stephan Oepen, Marco Kuhlmann, Yusuke Miyao, Daniel Zeman, Silvie Cinkov{\'a},
  Dan Flickinger, Jan Hajic, and Zdenka Uresova. 2015.
\newblock Semeval 2015 task 18: Broad-coverage semantic dependency parsing.
\newblock In \emph{Proceedings of the 9th International Workshop on Semantic
  Evaluation (SemEval 2015)}, pages 915--926.

\bibitem[{Oepen et~al.(2014)Oepen, Kuhlmann, Miyao, Zeman, Flickinger, Hajic,
  Ivanova, and Zhang}]{oepensemeval}
Stephan Oepen, Marco Kuhlmann, Yusuke Miyao, Daniel Zeman, Dan Flickinger, Jan
  Hajic, Angelina Ivanova, and Yi~Zhang. 2014.
\newblock Semeval 2014 task 8: Broad-coverage semantic dependency parsing.
\newblock \emph{SemEval 2014}.

\bibitem[{Owoputi et~al.(2013)Owoputi, O{'}Connor, Dyer, Gimpel, Schneider, and
  Smith}]{owoputi-etal-2013-improved}
Olutobi Owoputi, Brendan O{'}Connor, Chris Dyer, Kevin Gimpel, Nathan
  Schneider, and Noah~A. Smith. 2013.
\newblock \href {https://www.aclweb.org/anthology/N13-1039} {Improved
  part-of-speech tagging for online conversational text with word clusters}.
\newblock In \emph{Proceedings of the 2013 Conference of the North {A}merican
  Chapter of the Association for Computational Linguistics: Human Language
  Technologies}, pages 380--390, Atlanta, Georgia. Association for
  Computational Linguistics.

\bibitem[{Pennington et~al.(2014)Pennington, Socher, and
  Manning}]{pennington2014glove}
Jeffrey Pennington, Richard Socher, and Christopher Manning. 2014.
\newblock Glove: Global vectors for word representation.
\newblock In \emph{Proceedings of the 2014 conference on empirical methods in
  natural language processing (EMNLP)}, pages 1532--1543.

\bibitem[{Peters et~al.(2018)Peters, Neumann, Iyyer, Gardner, Clark, Lee, and
  Zettlemoyer}]{peters-etal-2018-deep}
Matthew Peters, Mark Neumann, Mohit Iyyer, Matt Gardner, Christopher Clark,
  Kenton Lee, and Luke Zettlemoyer. 2018.
\newblock \href {https://doi.org/10.18653/v1/N18-1202} {Deep contextualized
  word representations}.
\newblock In \emph{Proceedings of the 2018 Conference of the North {A}merican
  Chapter of the Association for Computational Linguistics: Human Language
  Technologies, Volume 1 (Long Papers)}, pages 2227--2237, New Orleans,
  Louisiana. Association for Computational Linguistics.

\bibitem[{Pham et~al.(2018{\natexlab{a}})Pham, Guan, Zoph, Le, and
  Dean}]{pmlr-v80-pham18a}
Hieu Pham, Melody Guan, Barret Zoph, Quoc Le, and Jeff Dean.
  2018{\natexlab{a}}.
\newblock Efficient neural architecture search via parameters sharing.
\newblock In \emph{International Conference on Machine Learning}, pages
  4095--4104.

\bibitem[{Pham et~al.(2018{\natexlab{b}})Pham, Guan, Zoph, Le, and
  Dean}]{pham2018efficient}
Hieu Pham, Melody~Y Guan, Barret Zoph, Quoc~V Le, and Jeff Dean.
  2018{\natexlab{b}}.
\newblock Efficient neural architecture search via parameter sharing.
\newblock In \emph{International Conference on Machine Learning}.

\bibitem[{Pires et~al.(2019)Pires, Schlinger, and
  Garrette}]{pires-etal-2019-multilingual}
Telmo Pires, Eva Schlinger, and Dan Garrette. 2019.
\newblock \href {https://doi.org/10.18653/v1/P19-1493} {How multilingual is
  multilingual {BERT}?}
\newblock In \emph{Proceedings of the 57th Annual Meeting of the Association
  for Computational Linguistics}, pages 4996--5001, Florence, Italy.
  Association for Computational Linguistics.

\bibitem[{Pontiki et~al.(2016)Pontiki, Galanis, Papageorgiou, Androutsopoulos,
  Manandhar, AL-Smadi, Al-Ayyoub, Zhao, Qin, De~Clercq, Hoste, Apidianaki,
  Tannier, Loukachevitch, Kotelnikov, Bel, Jim{\'e}nez-Zafra, and
  Eryi{\u{g}}it}]{pontiki-etal-2016-semeval}
Maria Pontiki, Dimitris Galanis, Haris Papageorgiou, Ion Androutsopoulos,
  Suresh Manandhar, Mohammad AL-Smadi, Mahmoud Al-Ayyoub, Yanyan Zhao, Bing
  Qin, Orph{\'e}e De~Clercq, V{\'e}ronique Hoste, Marianna Apidianaki, Xavier
  Tannier, Natalia Loukachevitch, Evgeniy Kotelnikov, Nuria Bel,
  Salud~Mar{\'\i}a Jim{\'e}nez-Zafra, and G{\"u}l{\c{s}}en Eryi{\u{g}}it. 2016.
\newblock \href {https://doi.org/10.18653/v1/S16-1002} {{S}em{E}val-2016 task
  5: Aspect based sentiment analysis}.
\newblock In \emph{Proceedings of the 10th International Workshop on Semantic
  Evaluation ({S}em{E}val-2016)}, pages 19--30, San Diego, California.
  Association for Computational Linguistics.

\bibitem[{Pontiki et~al.(2015)Pontiki, Galanis, Papageorgiou, Manandhar, and
  Androutsopoulos}]{pontiki-etal-2015-semeval}
Maria Pontiki, Dimitris Galanis, Haris Papageorgiou, Suresh Manandhar, and Ion
  Androutsopoulos. 2015.
\newblock \href {https://doi.org/10.18653/v1/S15-2082} {{S}em{E}val-2015 task
  12: Aspect based sentiment analysis}.
\newblock In \emph{Proceedings of the 9th International Workshop on Semantic
  Evaluation ({S}em{E}val 2015)}, pages 486--495, Denver, Colorado. Association
  for Computational Linguistics.

\bibitem[{Pontiki et~al.(2014)Pontiki, Galanis, Pavlopoulos, Papageorgiou,
  Androutsopoulos, and Manandhar}]{pontiki-etal-2014-semeval}
Maria Pontiki, Dimitris Galanis, John Pavlopoulos, Harris Papageorgiou, Ion
  Androutsopoulos, and Suresh Manandhar. 2014.
\newblock \href {https://doi.org/10.3115/v1/S14-2004} {{S}em{E}val-2014 task 4:
  Aspect based sentiment analysis}.
\newblock In \emph{Proceedings of the 8th International Workshop on Semantic
  Evaluation ({S}em{E}val 2014)}, pages 27--35, Dublin, Ireland. Association
  for Computational Linguistics.

\bibitem[{Real et~al.(2019)Real, Aggarwal, Huang, and Le}]{real2019regularized}
Esteban Real, Alok Aggarwal, Yanping Huang, and Quoc~V Le. 2019.
\newblock Regularized evolution for image classifier architecture search.
\newblock In \emph{Proceedings of the aaai conference on artificial
  intelligence}, volume~33, pages 4780--4789.

\bibitem[{Real et~al.(2017)Real, Moore, Selle, Saxena, Suematsu, Tan, Le, and
  Kurakin}]{real2017large}
Esteban Real, Sherry Moore, Andrew Selle, Saurabh Saxena, Yutaka~Leon Suematsu,
  Jie Tan, Quoc~V Le, and Alexey Kurakin. 2017.
\newblock Large-scale evolution of image classifiers.
\newblock In \emph{International Conference on Machine Learning}, pages
  2902--2911.

\bibitem[{Ritter et~al.(2011)Ritter, Clark, {Mausam}, and
  Etzioni}]{ritter-etal-2011-named}
Alan Ritter, Sam Clark, {Mausam}, and Oren Etzioni. 2011.
\newblock \href {https://www.aclweb.org/anthology/D11-1141} {Named entity
  recognition in tweets: An experimental study}.
\newblock In \emph{Proceedings of the 2011 Conference on Empirical Methods in
  Natural Language Processing}, pages 1524--1534, Edinburgh, Scotland, UK.
  Association for Computational Linguistics.

\bibitem[{Santos and Zadrozny(2014)}]{santos2014learning}
Cicero~D Santos and Bianca Zadrozny. 2014.
\newblock Learning character-level representations for part-of-speech tagging.
\newblock In \emph{Proceedings of the 31st international conference on machine
  learning (ICML-14)}, pages 1818--1826.

\bibitem[{Schuster et~al.(2019)Schuster, Ram, Barzilay, and
  Globerson}]{schuster-etal-2019-cross}
Tal Schuster, Ori Ram, Regina Barzilay, and Amir Globerson. 2019.
\newblock \href {https://doi.org/10.18653/v1/N19-1162} {Cross-lingual alignment
  of contextual word embeddings, with applications to zero-shot dependency
  parsing}.
\newblock In \emph{Proceedings of the 2019 Conference of the North {A}merican
  Chapter of the Association for Computational Linguistics: Human Language
  Technologies, Volume 1 (Long and Short Papers)}, pages 1599--1613,
  Minneapolis, Minnesota. Association for Computational Linguistics.

\bibitem[{So et~al.(2019)So, Liang, and Le}]{so2019evolved}
David~R So, Chen Liang, and Quoc~V Le. 2019.
\newblock The evolved transformer.
\newblock In \emph{International Conference on Machine Learning}.

\bibitem[{Stanley and Miikkulainen(2002)}]{stanley2002evolving}
Kenneth~O Stanley and Risto Miikkulainen. 2002.
\newblock Evolving neural networks through augmenting topologies.
\newblock \emph{Evolutionary computation}, 10(2):99--127.

\bibitem[{Strakov{\'a} et~al.(2019)Strakov{\'a}, Straka, and
  Hajic}]{strakova-etal-2019-neural}
Jana Strakov{\'a}, Milan Straka, and Jan Hajic. 2019.
\newblock \href {https://doi.org/10.18653/v1/P19-1527} {Neural architectures
  for nested {NER} through linearization}.
\newblock In \emph{Proceedings of the 57th Annual Meeting of the Association
  for Computational Linguistics}, pages 5326--5331, Florence, Italy.
  Association for Computational Linguistics.

\bibitem[{Suganuma et~al.(2017)Suganuma, Shirakawa, and
  Nagao}]{suganuma2017genetic}
Masanori Suganuma, Shinichi Shirakawa, and Tomoharu Nagao. 2017.
\newblock A genetic programming approach to designing convolutional neural
  network architectures.
\newblock In \emph{Proceedings of the genetic and evolutionary computation
  conference}, pages 497--504.

\bibitem[{Sundheim(1995)}]{Sundheim1995NamedET}
Beth~M. Sundheim. 1995.
\newblock Named entity task definition, version 2.1.
\newblock In \emph{Proceedings of the Sixth Message Understanding Conference},
  pages 319--332.

\bibitem[{Sutton and Barto(1992)}]{sutton2018reinforcement}
Richard~S Sutton and Andrew~G Barto. 1992.
\newblock \emph{Reinforcement learning: An introduction}.
\newblock MIT press.

\bibitem[{Szegedy et~al.(2016)Szegedy, Vanhoucke, Ioffe, Shlens, and
  Wojna}]{szegedy2016rethinking}
Christian Szegedy, Vincent Vanhoucke, Sergey Ioffe, Jon Shlens, and Zbigniew
  Wojna. 2016.
\newblock Rethinking the inception architecture for computer vision.
\newblock In \emph{Proceedings of the IEEE conference on computer vision and
  pattern recognition}, pages 2818--2826.

\bibitem[{Tesni{\`e}re(1959)}]{arrive1969elements}
Lucien Tesni{\`e}re. 1959.
\newblock {\'e}l{\'e}ments de syntaxe structurale.
\newblock \emph{Editions Klincksieck}.

\bibitem[{Tjong Kim~Sang(2002)}]{tjong-kim-sang-2002-introduction}
Erik~F. Tjong Kim~Sang. 2002.
\newblock \href {https://www.aclweb.org/anthology/W02-2024} {Introduction to
  the {C}o{NLL}-2002 shared task: Language-independent named entity
  recognition}.
\newblock In \emph{{COLING}-02: The 6th Conference on Natural Language Learning
  2002 ({C}o{NLL}-2002)}.

\bibitem[{Tjong Kim~Sang and
  Buchholz(2000)}]{tjong-kim-sang-buchholz-2000-introduction}
Erik~F. Tjong Kim~Sang and Sabine Buchholz. 2000.
\newblock \href {https://www.aclweb.org/anthology/W00-0726} {Introduction to
  the {C}o{NLL}-2000 shared task chunking}.
\newblock In \emph{Fourth Conference on Computational Natural Language Learning
  and the Second Learning Language in Logic Workshop}.

\bibitem[{Tjong Kim~Sang and
  De~Meulder(2003)}]{tjong-kim-sang-de-meulder-2003-introduction}
Erik~F. Tjong Kim~Sang and Fien De~Meulder. 2003.
\newblock \href {https://www.aclweb.org/anthology/W03-0419} {Introduction to
  the {C}o{NLL}-2003 shared task: Language-independent named entity
  recognition}.
\newblock In \emph{Proceedings of the Seventh Conference on Natural Language
  Learning at {HLT}-{NAACL} 2003}, pages 142--147.

\bibitem[{Vaswani et~al.(2017)Vaswani, Shazeer, Parmar, Uszkoreit, Jones,
  Gomez, Kaiser, and Polosukhin}]{vaswani2017attention}
Ashish Vaswani, Noam Shazeer, Niki Parmar, Jakob Uszkoreit, Llion Jones,
  Aidan~N Gomez, {\L}ukasz Kaiser, and Illia Polosukhin. 2017.
\newblock Attention is all you need.
\newblock In \emph{Advances in neural information processing systems}, pages
  5998--6008.

\bibitem[{Wang et~al.(2019)Wang, Huang, and Tu}]{wang-etal-2019-second}
Xinyu Wang, Jingxian Huang, and Kewei Tu. 2019.
\newblock \href {https://doi.org/10.18653/v1/P19-1454} {Second-order semantic
  dependency parsing with end-to-end neural networks}.
\newblock In \emph{Proceedings of the 57th Annual Meeting of the Association
  for Computational Linguistics}, pages 4609--4618, Florence, Italy.
  Association for Computational Linguistics.

\bibitem[{Wang et~al.(2020{\natexlab{a}})Wang, Jiang, Bach, Wang, Huang, and
  Tu}]{wang-etal-2020-structure}
Xinyu Wang, Yong Jiang, Nguyen Bach, Tao Wang, Fei Huang, and Kewei Tu.
  2020{\natexlab{a}}.
\newblock \href {https://doi.org/10.18653/v1/2020.acl-main.304}
  {Structure-level knowledge distillation for multilingual sequence labeling}.
\newblock In \emph{Proceedings of the 58th Annual Meeting of the Association
  for Computational Linguistics}, pages 3317--3330, Online. Association for
  Computational Linguistics.

\bibitem[{Wang et~al.(2021{\natexlab{a}})Wang, Jiang, Bach, Wang, Huang, Huang,
  and Tu}]{wang2021improving}
Xinyu Wang, Yong Jiang, Nguyen Bach, Tao Wang, Zhongqiang Huang, Fei Huang, and
  Kewei Tu. 2021{\natexlab{a}}.
\newblock {{Improving Named Entity Recognition by External Context Retrieving
  and Cooperative Learning}}.
\newblock In \emph{{the Joint Conference of the 59th Annual Meeting of the
  Association for Computational Linguistics and the 11th International Joint
  Conference on Natural Language Processing (\textbf{ACL-IJCNLP 2021})}}.
  Association for Computational Linguistics.

\bibitem[{Wang et~al.(2020{\natexlab{b}})Wang, Jiang, Bach, Wang, Zhongqiang,
  Huang, and Tu}]{wang-etal-2020-more}
Xinyu Wang, Yong Jiang, Nguyen Bach, Tao Wang, Huang Zhongqiang, Fei Huang, and
  Kewei Tu. 2020{\natexlab{b}}.
\newblock More embeddings, better sequence labelers?
\newblock In \emph{Findings of EMNLP}, Online.

\bibitem[{Wang et~al.(2021{\natexlab{b}})Wang, Jiang, Yan, Jia, Bach, Wang,
  Huang, Huang, and Tu}]{wang2020structural}
Xinyu Wang, Yong Jiang, Zhaohui Yan, Zixia Jia, Nguyen Bach, Tao Wang,
  Zhongqiang Huang, Fei Huang, and Kewei Tu. 2021{\natexlab{b}}.
\newblock {{Structural Knowledge Distillation: Tractably Distilling Information
  for Structured Predictor}}.
\newblock In \emph{{the Joint Conference of the 59th Annual Meeting of the
  Association for Computational Linguistics and the 11th International Joint
  Conference on Natural Language Processing (\textbf{ACL-IJCNLP 2021})}}.
  Association for Computational Linguistics.

\bibitem[{Wang and Tu(2020)}]{wang-tu-2020-second}
Xinyu Wang and Kewei Tu. 2020.
\newblock \href {https://www.aclweb.org/anthology/2020.aacl-main.12}
  {Second-order neural dependency parsing with message passing and end-to-end
  training}.
\newblock In \emph{Proceedings of the 1st Conference of the Asia-Pacific
  Chapter of the Association for Computational Linguistics and the 10th
  International Joint Conference on Natural Language Processing}, pages 93--99,
  Suzhou, China. Association for Computational Linguistics.

\bibitem[{Wei et~al.(2020)Wei, Hong, Zou, Cheng, and Yao}]{wei-etal-2020-dont}
Zhenkai Wei, Yu~Hong, Bowei Zou, Meng Cheng, and Jianmin Yao. 2020.
\newblock \href {https://doi.org/10.18653/v1/2020.acl-main.339} {Don{'}t
  eclipse your arts due to small discrepancies: Boundary repositioning with a
  pointer network for aspect extraction}.
\newblock In \emph{Proceedings of the 58th Annual Meeting of the Association
  for Computational Linguistics}, pages 3678--3684, Online. Association for
  Computational Linguistics.

\bibitem[{Williams(1992)}]{williams1992simple}
Ronald~J Williams. 1992.
\newblock Simple statistical gradient-following algorithms for connectionist
  reinforcement learning.
\newblock \emph{Machine learning}, 8(3-4):229--256.

\bibitem[{Wistuba(2018)}]{wistuba2018deep}
Martin Wistuba. 2018.
\newblock Deep learning architecture search by neuro-cell-based evolution with
  function-preserving mutations.
\newblock In \emph{Joint European Conference on Machine Learning and Knowledge
  Discovery in Databases}, pages 243--258. Springer.

\bibitem[{Wu and Dredze(2019)}]{wu-dredze-2019-beto}
Shijie Wu and Mark Dredze. 2019.
\newblock \href {https://doi.org/10.18653/v1/D19-1077} {Beto, bentz, becas: The
  surprising cross-lingual effectiveness of {BERT}}.
\newblock In \emph{Proceedings of the 2019 Conference on Empirical Methods in
  Natural Language Processing and the 9th International Joint Conference on
  Natural Language Processing (EMNLP-IJCNLP)}, pages 833--844, Hong Kong,
  China. Association for Computational Linguistics.

\bibitem[{{Xie} and {Yuille}(2017)}]{xie2017genetic}
L.~{Xie} and A.~{Yuille}. 2017.
\newblock Genetic cnn.
\newblock In \emph{2017 IEEE International Conference on Computer Vision
  (ICCV)}, pages 1388--1397.

\bibitem[{Xu et~al.(2019)Xu, Liu, Shu, and Yu}]{xu-etal-2019-bert}
Hu~Xu, Bing Liu, Lei Shu, and Philip Yu. 2019.
\newblock \href {https://doi.org/10.18653/v1/N19-1242} {{BERT} post-training
  for review reading comprehension and aspect-based sentiment analysis}.
\newblock In \emph{Proceedings of the 2019 Conference of the North {A}merican
  Chapter of the Association for Computational Linguistics: Human Language
  Technologies, Volume 1 (Long and Short Papers)}, pages 2324--2335,
  Minneapolis, Minnesota. Association for Computational Linguistics.

\bibitem[{Xu et~al.(2018)Xu, Liu, Shu, and Yu}]{xu-etal-2018-double}
Hu~Xu, Bing Liu, Lei Shu, and Philip~S. Yu. 2018.
\newblock \href {https://doi.org/10.18653/v1/P18-2094} {Double embeddings and
  {CNN}-based sequence labeling for aspect extraction}.
\newblock In \emph{Proceedings of the 56th Annual Meeting of the Association
  for Computational Linguistics (Volume 2: Short Papers)}, pages 592--598,
  Melbourne, Australia. Association for Computational Linguistics.

\bibitem[{Yamada et~al.(2020)Yamada, Asai, Shindo, Takeda, and
  Matsumoto}]{yamada-etal-2020-luke}
Ikuya Yamada, Akari Asai, Hiroyuki Shindo, Hideaki Takeda, and Yuji Matsumoto.
  2020.
\newblock \href {https://doi.org/10.18653/v1/2020.emnlp-main.523} {{LUKE}: Deep
  contextualized entity representations with entity-aware self-attention}.
\newblock In \emph{Proceedings of the 2020 Conference on Empirical Methods in
  Natural Language Processing (EMNLP)}, pages 6442--6454, Online. Association
  for Computational Linguistics.

\bibitem[{Yang et~al.(2019)Yang, Dai, Yang, Carbonell, Salakhutdinov, and
  Le}]{yang2019xlnet}
Zhilin Yang, Zihang Dai, Yiming Yang, Jaime Carbonell, Russ~R Salakhutdinov,
  and Quoc~V Le. 2019.
\newblock Xlnet: Generalized autoregressive pretraining for language
  understanding.
\newblock In \emph{Advances in neural information processing systems}, pages
  5753--5763.

\bibitem[{Yu et~al.(2020)Yu, Bohnet, and Poesio}]{yu-etal-2020-named}
Juntao Yu, Bernd Bohnet, and Massimo Poesio. 2020.
\newblock \href {https://doi.org/10.18653/v1/2020.acl-main.577} {Named entity
  recognition as dependency parsing}.
\newblock In \emph{Proceedings of the 58th Annual Meeting of the Association
  for Computational Linguistics}, pages 6470--6476, Online. Association for
  Computational Linguistics.

\bibitem[{Zhang et~al.(2020)Zhang, Li, and Zhang}]{zhang-etal-2020-efficient}
Yu~Zhang, Zhenghua Li, and Min Zhang. 2020.
\newblock \href {https://doi.org/10.18653/v1/2020.acl-main.302} {Efficient
  second-order {T}ree{CRF} for neural dependency parsing}.
\newblock In \emph{Proceedings of the 58th Annual Meeting of the Association
  for Computational Linguistics}, pages 3295--3305, Online. Association for
  Computational Linguistics.

\bibitem[{Zhong et~al.(2018)Zhong, Yan, Wu, Shao, and Liu}]{zhong2018practical}
Zhao Zhong, Junjie Yan, Wei Wu, Jing Shao, and Cheng-Lin Liu. 2018.
\newblock Practical block-wise neural network architecture generation.
\newblock In \emph{Proceedings of the IEEE conference on computer vision and
  pattern recognition}, pages 2423--2432.

\bibitem[{Zhou and Zhao(2019)}]{zhou-zhao-2019-head}
Junru Zhou and Hai Zhao. 2019.
\newblock \href {https://doi.org/10.18653/v1/P19-1230} {{H}ead-{D}riven
  {P}hrase {S}tructure {G}rammar parsing on {P}enn {T}reebank}.
\newblock In \emph{Proceedings of the 57th Annual Meeting of the Association
  for Computational Linguistics}, pages 2396--2408, Florence, Italy.
  Association for Computational Linguistics.

\bibitem[{Zhu et~al.(2020)Zhu, Wang, Qiu, Ni, and Xie}]{zhu2020autotrans}
Wei Zhu, Xiaoling Wang, Xipeng Qiu, Yuan Ni, and Guotong Xie. 2020.
\newblock Autotrans: Automating transformer design via reinforced architecture
  search.
\newblock \emph{arXiv preprint arXiv:2009.02070}.

\bibitem[{Zoph and Le(2017)}]{zoph2016neural}
Barret Zoph and Quoc~V Le. 2017.
\newblock Neural architecture search with reinforcement learning.
\newblock In \emph{International Conference on Learning Representations}.

\bibitem[{Zoph et~al.(2018)Zoph, Vasudevan, Shlens, and Le}]{zoph2018learning}
Barret Zoph, Vijay Vasudevan, Jonathon Shlens, and Quoc~V Le. 2018.
\newblock Learning transferable architectures for scalable image recognition.
\newblock In \emph{Proceedings of the IEEE conference on computer vision and
  pattern recognition}, pages 8697--8710.

\end{thebibliography}
\appendix
\section{Detailed Configurations}
\label{app:detail}
\paragraph{Evaluation}
To evaluate our models, We use F1 score to evaluate NER, Chunking and AE, use accuracy to evaluate POS Tagging, use unlabeled attachment score (UAS) and labeled attachment score (LAS) to evaluate DP, and use labeled F1 score to evaluate SDP. 

\paragraph{Task Models and Controller}
\label{sec:models}
For sequence-structured tasks (i.e., NER, POS tagging, chunking, aspect extraction), we use a batch size of $32$ sentences and an SGD optimizer with a learning rate of $0.1$. We anneal the learning rate by $0.5$ when there is no accuracy improvement on the development set for 5 epochs. We set the maximum training epoch to $150$. For graph-structured tasks (i.e., DP and SDP), we use Adam \citep{kingma2014adam} to optimize the model with a learning rate of $0.002$. We anneal the learning rate by $0.75$ for every $5000$ iterations following \citet{dozat2016deep}. We set the maximum training epoch to $300$. For DP, we run the maximum spanning tree \cite{mcdonald-etal-2005-non} algorithm to output valid trees in testing. We fix the hyper-parameters of the task models. 

We tune the learning rate for the controller among $\{0.1,0.2,0.3,0.4,0.5\}$ and the discount factor among $\{0.1,0.3,0.5,0.7,0.9\}$ on the same dataset in Section \ref{sec:ablation}. We search for the hyper-parameter through grid search and find a learning rate of $0.1$ and a discount factor of $0.5$ performs the best on the development set. The controller's parameters are initialized to all $0$ so that each candidate is selected evenly in the first two time steps. 
We use Stochastic Gradient Descent (SGD) to optimize the controller. 
The training time depends on the task and dataset size. Take the CoNLL English NER dataset as an example. It takes $45$ GPU hours to train the controller for $30$ steps on a single Tesla P100 GPU, which is an acceptable training time in practice.

\paragraph{Sources of Embeddings}
\label{sec:embed}
The sources of the embeddings that we used are listed in Table \ref{tab:embeddings}. 
\begin{table*}[!ht]
\tiny
\centering
\begin{tabular}{l|l|l}
\hlineB{4}
\bf \textsc{Embedding} & \bf \textsc{Resource} & \bf \textsc{URL}\\
\hline
GloVe & \citet{pennington2014glove} & \url{nlp.stanford.edu/projects/glove}\\
fastText & \citet{bojanowski2017enriching} & \url{github.com/facebookresearch/fastText}\\
ELMo & \citet{peters-etal-2018-deep} & \url{github.com/allenai/allennlp}\\
ELMo (Other languages) & \citet{schuster-etal-2019-cross} & \url{github.com/TalSchuster/CrossLingualContextualEmb}\\
BERT & \citet{devlin-etal-2019-bert} & \url{huggingface.co/bert-base-cased}\\
M-BERT & \citet{devlin-etal-2019-bert} & \url{huggingface.co/bert-base-multilingual-cased}\\
BERT (Dutch) & wietsedv & \url{huggingface.co/wietsedv/bert-base-dutch-cased}\\
BERT (German) & dbmdz & \url{huggingface.co/bert-base-german-dbmdz-cased}\\
BERT (Spanish) & dccuchile & \url{huggingface.co/dccuchile/bert-base-spanish-wwm-cased}\\
BERT (Turkish) & dbmdz & \url{huggingface.co/dbmdz/bert-base-turkish-cased}\\
XLM-R & \citet{conneau-etal-2020-unsupervised} & \url{huggingface.co/xlm-roberta-large}\\
RoBERTa & \citet{liu2019roberta} & \url{huggingface.co/roberta-large}\\
XLNet & \citet{yang2019xlnet} & \url{huggingface.co/xlnet-large-cased}\\
\hlineB{4}
\end{tabular}
\caption{The embeddings we used in our experiments. The URL is where we downloaded the embeddings. }
\label{tab:embeddings}
\end{table*}

\begin{table}[t!]
\small
\centering
\setlength\tabcolsep{4pt}
\begin{tabular}{l|ccccc}
\hlineB{4}
  & de   & de$_\text{06}$ & en   & es   & nl \\
\hline\hline
All+sent & 86.8 & 90.1 & 93.3 & 90.0 & 94.4 \\
ACE+sent & 87.1 & 90.5 & 93.6 & 92.4 & 94.6 \\
\hline
BERT (\citeyear{devlin-etal-2019-bert})    &  -    & - & 92.8 & -    & -\\
\citet{akbik-etal-2019-pooled}     &  -    & 88.3 & 93.2 & -    & 90.4\\
\citet{yu-etal-2020-named} & 86.4 & 90.3 & 93.5 & 90.3 & 94.7 \\
\citet{yamada-etal-2020-luke} & - & - & 94.3 & - & - \\ 
\citet{luoma-pyysalo-2020-exploring} & 87.3 & - & 93.7 & 88.3 & 93.5 \\
\citet{wang2021improving} & - & - & 93.9 & - & -  \\
All+doc & 87.5 & 90.8 & 94.0 & 90.7 & 93.7 \\
ACE+doc &  \textbf{88.3} & \textbf{91.7} & \textbf{94.6} & \textbf{95.9} & \textbf{95.7}\\
\hlineB{4}
\end{tabular}
\caption{Comparison of models with and without document contexts on NER. +sent/+doc: models with sentence-/document-level embeddings.}
\label{tab:document}
\end{table}

\section{Additional Analysis}

\subsection{Document-Level and Sentence-Level Representations}
Recently, models with document-level word representations extracted from transformer-based embeddings significantly outperform models with sentence-level word representations in NER \citep{devlin-etal-2019-bert,yu-etal-2020-named,yamada-etal-2020-luke}. However, there are a lot of application scenarios that document contexts are unavailable. We replace the document-level word representations from transformer-based embeddings (i.e., XLM-R and BERT embeddings) with the sentence-level word representations. Results are shown in Table \ref{tab:document}. We report the test results of \texttt{All} to show how the gap between ACE and \texttt{All} changes with different kinds of representations. We report the test accuracy of the models with the highest development accuracy following \citet{yamada-etal-2020-luke} for a fair comparison. Empirical results show that the document-level representations can significantly improve the accuracy of ACE. Comparing with models with sentence-level representations, the averaged accuracy gap between ACE and \texttt{All} is enhanced from 0.7 to 1.7 with document-level representations, which shows that the advantage of ACE becomes stronger with document-level representations.

\subsection{Fine-tuned Models Versus ACE}
\label{sec:finetune}
To fine-tune the embeddings, we use AdamW \citep{loshchilov2018decoupled} optimizer with a learning rate of $5\times 10^{-6}$ and trained the contextualized embeddings with the task for 10 epochs. We use a batch size of 32 for BERT, M-BERT and use a batch size of 4 for XLM-R, RoBERTa and XLNet. A comparison between ACE and the fine-tuned embeddings that we used in ACE is shown in Table \ref{tab:ft_vs_ace_0}, \ref{tab:ft_vs_ace_1}. Results show that ACE can further improve the accuracy of fine-tuned models. 

\subsection{Retraining}
Most of the work \citep{zoph2016neural,zoph2018learning,pham2018efficient,so2019evolved,zhu2020autotrans} in NAS retrains the searched neural architecture from scratch so that the hyper-parameters of the searched model can be modified or trained on larger datasets. To show whether our searched embedding concatenation is helpful to the task, we retrain the task model with the embedding concatenations on the same dataset from scratch. For the experiment, we use the same dataset settings as in Section \ref{sec:exp:base}. We train the searched embedding concatenation of each run from ACE 3 times (therefore, 9 runs for each dataset). 

Table \ref{tab:direct} shows the comparison between retrained models with the searched embedding concatenation from ACE and \texttt{All}. The results show that the retrained models are competitive with ACE in SDP and in chunking. However, in another three tasks, the retrained models perform inferior to ACE. The possible reason is that the model at each step is initialized by the trained model of previous step. The retrained models outperform \texttt{All} in all tasks, which shows the effectiveness of the searched embedding concatenations.

\subsection{Effect of Embeddings in the Searched Embedding Concatenations}
There is no clear conclusion on what concatenation of embeddings is helpful to most of the tasks. 
We analyze the best searched embedding concatenations by ACE over different structured outputs, semantic/syntactic type, and monolingual/multilingual tasks. 
The percentage of each embedding selected by the best concatenations from all experiments of ACE are shown in Table \ref{tab:selection}. 
The best embedding concatenation varies over the output structure, syntactic/semantic level of understanding, and the language. 
The experimental results show that it is essential to select embeddings for each kind of task separately. 
However, we also find that the embeddings are strong in specific settings. In comparison to the sequence-structured and graph-structured tasks, we find that M-BERT and ELMo are only frequently selected in sequence-structured tasks while XLM-R embeddings are always selected in graph-structured tasks. 
For Flair embeddings, the forward and backward model are evenly selected. 
We suspect one direction of Flair embeddings is strong enough. 
Therefore concatenating the embeddings from two directions together cannot further improve the accuracy. 
For non-contextualized embeddings, pretrained word embeddings are frequently selected in sequence-structured tasks, and character embeddings are not.
When we dig deeper into the semantic and syntactic type of these two structured outputs, we find that in all best concatenations, BERT embeddings are selected in all syntactic sequence-structured tasks, and Flair, M-Flair, word, and XLM-R embeddings are selected in syntactic graph-structured tasks. 
In multilingual tasks, all best concatenations in multilingual NER tasks select M-BERT embeddings while M-BERT is rarely selected in multilingual AE tasks. 
The monolingual Flair embeddings are always selected in NER tasks, and XLM-R is more frequently selected in multilingual tasks than monolingual sequence-structured tasks (\textbf{SS}).

\begin{table*}[!ht]
\small
\centering
\begin{tabular}{l||ccccc|ccc}
\hlineB{4}
& \multicolumn{5}{c|}{NER}  & \multicolumn{3}{c}{POS}\\
\hhline{~||-----|---}
                & de   & de (Revised) & en   & es   & nl   & Ritter & ARK  & TB-v2 \\
\hline\hline
BERT+Fine-tune            & 76.9 & 79.4 & 89.2 & 83.3 & 83.8 & 91.2   & 91.7 & 94.4  \\
MBERT+Fine-tune           & 81.6 & 86.7 & 92.0 & 87.1 & 87.2 & 90.8   & 91.5 & 93.9  \\
XLM-R+Fine-tune & 87.7    & 91.4 & 94.1 & 89.3    & 95.3    & 92.3   & 93.7 & 95.4 \\
RoBERTa+Fine-tune & -    & - & 93.9 & -    & -    & 92.0   & 93.9 & 95.4 \\
XLNET+Fine-tune & -    & - & 93.6 & -    & -    & 88.4   & 92.4 & 94.4 \\
\hline
ACE+Fine-tune & \textbf{88.3} & \textbf{91.7} & \textbf{94.6} & \textbf{95.9} & \textbf{95.7} & \textbf{93.4}   & \textbf{94.4} & \textbf{95.8} \\
\hlineB{4}
\end{tabular}
\caption{A comparison between ACE and the fine-tuned embeddings that are used in ACE for NER and POS tagging.}
\label{tab:ft_vs_ace_0}
\end{table*}

\begin{table*}[!ht]
\small
\centering
\setlength\tabcolsep{4pt}
\begin{tabular}{l||c|cccccccc}
\hlineB{4}
 & \multicolumn{1}{c|}{Chunk} & \multicolumn{8}{c}{AE}  \\
\hhline{~||------|---}
                  & CoNLL 2000 & 14Lap & 14Res & 15Res & 16Res & es   & nl   & ru   & tr \\
\hline\hline
BERT+Fine-tune & 96.7 & 81.2  & 87.7  & 71.8  & 73.9  & 76.9 & 73.1 & 64.3 & 75.6   \\
MBERT+Fine-tune  & 96.6 & 83.5  & 85.0  & 69.5  & 73.6  & 74.5 & 72.6 & 71.6 & 58.8  \\
XLM-R+Fine-tune  & 97.0 & 85.9  & 90.5  & 76.4  & 78.9  & 77.0 & 77.6 & 77.7 & 74.1  \\
RoBERTa+Fine-tune  & 97.2 & 83.9  & 90.2  & 78.5  & 80.7  & - & - & - & -  \\
XLNET+Fine-tune  & 97.1 & 84.5  & 88.9  & 72.8  & 73.4  & - & - & - & -  \\
\hline
ACE+Fine-tune  &  \textbf{97.3}    & \textbf{87.4}   & \textbf{92.0} & \textbf{80.3}  & \textbf{81.3}  & \textbf{79.9} & \textbf{80.5} & \textbf{79.4} & \textbf{81.9} \\
\hlineB{4}
\end{tabular}
\caption{A comparison between ACE and the fine-tuned embeddings we used in ACE for chunking and AE.}
\label{tab:ft_vs_ace_1}
\end{table*}

\begin{table*}[!ht]
\small
\centering
\begin{tabular}{l||cc|ccccccccc}
\hlineB{4}
                & \multicolumn{2}{c|}{DP}  & \multicolumn{6}{c}{SDP}\\
\hhline{~||--|------}
                & \multicolumn{2}{c|}{PTB} & \multicolumn{2}{c}{DM} & \multicolumn{2}{c}{PAS} & \multicolumn{2}{c}{PSD} \\
                & UAS         & LAS       & ID         & OOD       & ID         & OOD        & ID         & OOD        \\
\hline\hline
BERT+Fine-tune            & 96.6        & 95.1      & 94.4       & 91.4      & 94.4       & 93.0       & 82.0       & 81.3       \\
MBERT+Fine-tune           & 96.5        & 94.9      & 93.9       & 90.4      & 93.9       & 92.1       & 81.2       & 80.0       \\
XLM-R+Fine-tune & 96.7        & 95.4      & 94.2       & 90.4      & 94.6       & 93.2       & 82.9       & 81.7       \\
RoBERTa+Fine-tune & 96.9        & 95.6      & 93.0       & 89.3      & 94.3       & 92.8       & 82.0       & 80.6       \\
XLNET+Fine-tune           & 97.0        & 95.6      & 94.2       & 90.6      & 94.8       & 93.4       & 82.7       & 81.8\\
\hline
ACE+Fine-tune  & \textbf{97.2}    & \textbf{95.7} & \textbf{95.6} & \textbf{92.6}    & \textbf{95.8} & \textbf{94.6}     & \textbf{83.8} & \textbf{83.4}    \\
\hlineB{4}
\end{tabular}
\caption{A comparison between ACE and the fine-tuned embeddings that are used in ACE for DP and SDP.}
\label{tab:ft_vs_ace_2}
\end{table*}

\begin{table*}[!ht]
\small
\centering
\begin{tabular}{l||cccccccc}
\hlineB{4}
             & NER  & POS  & Chunk & AE   & DP-UAS & DP-LAS & SDP-ID  & SDP-OOD \\
\hline
\hline
All          & 92.4 & 90.6 & 96.7  & 73.2 & 96.7   & 95.1   & 94.3 & 90.8    \\
Retrain & 92.6 & 90.8 & \textbf{96.8}  & 73.6 & 96.8   &95.2   & \textbf{94.5} & \textbf{90.9}    \\
ACE          & \textbf{93.0} & \textbf{91.7} & \textbf{96.8}  & \textbf{75.6} & \textbf{96.9}   & \textbf{95.3}   & \textbf{94.5} & \textbf{90.9}   \\
\hlineB{4}
\end{tabular}
\caption{A comparison among retrained models, All and ACE. We use the one dataset for each task.}
\label{tab:direct}
\end{table*}

\begin{table*}[t!]
\small
\centering
\setlength\tabcolsep{2.2pt}
\begin{tabular}{l||cccccccccccc}
\hlineB{4}
          & BERT & M-BERT & Char & ELMo & F    & F-bw & F-fw & MF   & MF-bw & MF-fw & Word & XLM-R \\
\hline\hline
SS        & 0.81 & 0.74   & 0.37 & 0.85 & 0.70 & 0.48 & 0.59 & 0.78 & 0.59  & 0.41  & 0.81 & 0.70  \\
GS      & 0.75 & 0.17   & 0.50 & 0.25 & 0.83 & 0.75 & 0.42 & 0.83 & 0.58  & 0.58  & 0.50 & 1.00  \\
\hline
Sem. SS   & 0.67 & 0.73   & 0.40 & 0.80 & 0.60 & 0.40 & 0.53 & 0.87 & 0.60  & 0.53  & 0.80 & 0.60  \\
Syn. SS   & 1.00 & 0.75   & 0.33 & 0.92 & 0.83 & 0.58 & 0.67 & 0.67 & 0.58  & 0.25  & 0.83 & 0.83  \\
\hline
Sem. GS & 0.78 & 0.22   & 0.67 & 0.33 & 0.78 & 0.67 & 0.56 & 0.78 & 0.56  & 0.67  & 0.33 & 1.00  \\
Syn. GS & 0.67 & 0.00   & 0.00 & 0.00 & 1.00 & 1.00 & 0.00 & 1.00 & 0.67  & 0.33  & 1.00 & 1.00  \\
\hline
M-NER     & 0.67 & 1.00   & 0.56 & 0.83 & 1.00 & 0.78 & 1.00 & 0.89 & 0.78  & 0.44  & 0.78 & 0.89  \\
M-AE      & 1.00 & 0.33   & 0.75 & 0.33 & 0.58 & 0.42 & 0.42 & 0.75 & 0.25  & 0.75  & 0.50 & 0.92  \\
\hlineB{4}
\end{tabular}
\caption{The percentage of each embedding candidate selected in the best concatenations from ACE. \textbf{F} and  \textbf{MF} are monolingual and multilingual Flair embeddings. We count these two embeddings are selected if one of the forward/backward (\textbf{fw}/\textbf{bw}) direction of Flair is selected in the concatenation. We count the \textbf{Word} embedding is selected if one of the fastText/GloVe embeddings is selected. \textbf{SS}: sequence-structured tasks. \textbf{GS}: graph-structured tasks. \textbf{Sem.}: Semantic-level tasks. \textbf{Syn.}: Syntactic-level tasks. \textbf{M-NER}: Multilingual NER tasks. \textbf{M-AE}: Multilingual AE tasks. We only use English datasets in \textbf{SS} and \textbf{GS}. English datasets are removed for \textbf{M-NER} and \textbf{M-AE}.}
\label{tab:selection}
\end{table*}


\end{document}